\documentclass{article}

\usepackage[preprint]{neurips_2024}

\usepackage[utf8]{inputenc} %
\usepackage[T1]{fontenc}    %
\usepackage{hyperref}       %
\usepackage{url}            %
\usepackage{booktabs}       %
\usepackage{amsfonts}       %
\usepackage{microtype}      %
\usepackage{xcolor}         %
\usepackage{amsmath}
\usepackage{amssymb}
\usepackage{mathtools}
\usepackage{amsthm}
\usepackage{algorithm}
\usepackage{algorithmic}

\usepackage[toc,page,header]{appendix}
\usepackage{minitoc}

\newcommand{\observed}{X}

\newcommand{\pert}{\delta}

\usepackage{shortex}

\makeatletter
\def\paragraph{\@startsection{paragraph}{4}{\z@}{0ex}{-1em}{\normalsize\bf}}
\def\subparagraph{\@startsection{subparagraph}{5}{\z@}{0ex}{-1em}{\normalsize\sc}}
\makeatother

\title{Automated Discovery of Pairwise Interactions from Unstructured Data}

\author{%
    Zuheng (David) Xu \thanks{Work done during an internship at Valence Labs} \\
  Department of Statistics\\
  University of British Columbia\\
  Vancouver, Canada\\
  \texttt{zuheng.xu@stat.ubc.ca} \\
  \And
   Moksh Jain$^*$ \\
  Department of Computer Science\\
  Mila / Université de Montréal\\
  Montreal, Canada\\
  \texttt{moksh.jain@mila.quebec} \\
    \And
  Ali Denton\\
  Valence Labs\\
  Montreal, Canada\\
  \texttt{ali@valencelabs.com} \\
    \And
  Shawn Whitfield\\
  Valence Labs\\
  Montreal, Canada\\
  \texttt{shawn@valencelabs.com} \\
    \And
       Aniket Didolkar$^*$ \\
  Department of Computer Science\\
  Mila / Université de Montréal\\
  Montreal, Canada\\
  \texttt{aniket.didolkar@mila.quebec} \\
    \And
      Berton Earnshaw \\
  Valence Labs \& Recursion\\
  Salt Lake City, USA\\
  \texttt{berton@valencelabs.com} \\
    \And
      Jason Hartford \\
  Valence Labs\\
  London, UK\\
  \texttt{jason@valencelabs.com} 
}

\begin{document}

\maketitle

\begin{abstract}
Pairwise interactions between perturbations to a system can provide evidence for the causal dependencies of the underlying underlying mechanisms of a system. When observations are low dimensional, hand crafted measurements, detecting interactions amounts to simple statistical tests, but it is not obvious how to detect interactions between perturbations affecting latent variables. %
We derive two interaction tests that are based on pairwise interventions, and show how these tests can be integrated into an active learning pipeline to efficiently discover pairwise interactions between perturbations.
We illustrate the value of these tests in the context of biology, where
pairwise perturbation experiments are frequently used to reveal interactions
that are not observable from any single perturbation. 
Our tests can be run on unstructured data, such as
the pixels in an image, which enables a more general notion of interaction than
typical cell viability experiments, and can be run on cheaper experimental assays. 
We validate on several synthetic and real biological experiments that our tests are able to identify interacting pairs effectively. 
We evaluate our approach on a real biological experiment where we knocked out
50 pairs of genes and measured the effect with microscopy images. We show that
we are able to recover significantly more known biological interactions than
random search and standard active learning baselines. 
\end{abstract}

\doparttoc %
\faketableofcontents %

\section{Introduction}

\begin{figure}
    \centering
    \includegraphics[width=\textwidth]{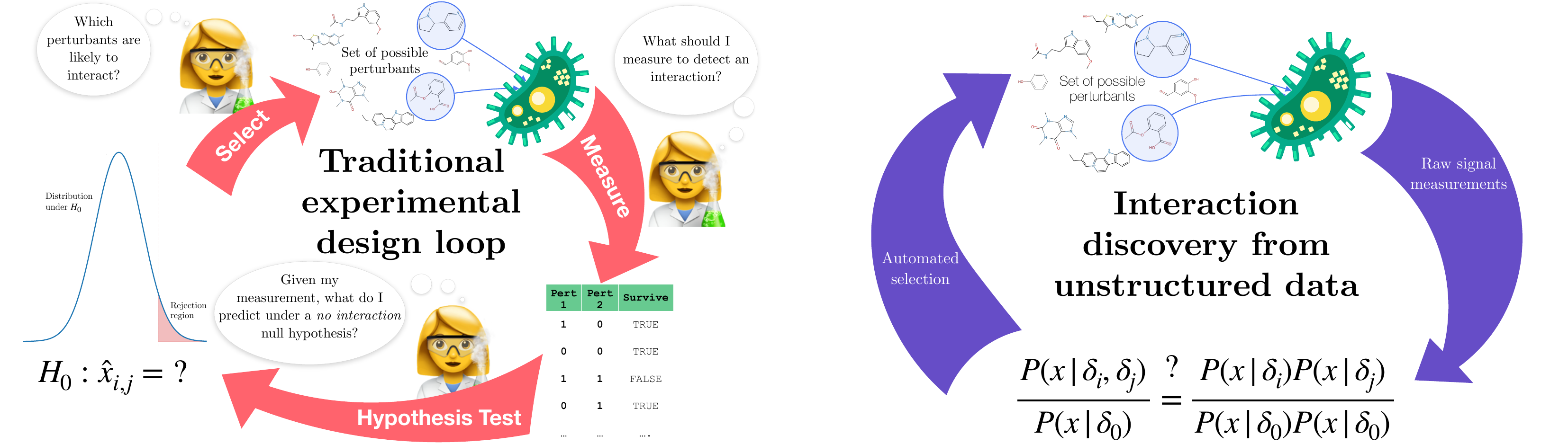}
    \caption{The traditional experimental design loop involves a number of human expert decisions: the expert need to \emph{measure} a carefully chosen feature of the experimental outcome (e.g. did a cell survive a perturbation?), formulate a prediction for the behaviour of this feature under the null hypothesis to test for interactions, and select interacting pairs for the combinatorial space of possible perturbations. Our approach enables automated selection of perturbants and testing for interactions directly from raw signal data (e.g. images of cells under a microscope).} 
    \label{fig:enter-label}
\end{figure}

Across the sciences, measurement of pairwise interaction between perturbations often reveals the existence of underlying mechanisms that single perturbations cannot. For example, quantum entanglement experiments support the counterintuitive predictions of quantum physics by demonstrating that entangled particles' spins which we would expect (under classical laws) to be independent, are in fact perfect anti-correlated once we make a measurement.
In economics, people may be more (or less) willing to pay for goods when presented in a bundle than they are to pay for each good in isolation, which reveals complements (or substitutes) in the underlying consumer preferences.
And in biology---the application area on which we focus---the concept of \emph{synthetic lethality} \citep{Nijman2011} occurs when the simultaneous perturbation of two genes results in cell death while individual perturbations do not. Synthetic lethality reveals that the genes' associated proteins play redundant roles in the underlying cellular mechanisms. 

In order to demonstrate a pairwise interaction a scientist (the human expert) has to carefully design three steps of the experiment: 
\begin{enumerate}
\item \textbf{Measurement:} the human expert to decides on what specific properties of the system to measure in order to reveal the interaction. %
For example, measuring a particle's spin may reveal entanglement, while the mass of a particle will not; similarly measuring cell viability may reveal synthetic lethality, while measuring the cell's colour will not. 
\item \textbf{Hypothesis testing:} an interaction is, by definition, a deviation from the effect that we expect under some null hypothesis which assumes each perturbation acts independently. The expert needs to specify the expected outcome assuming independence and %
compare this prediction to the actual outcome.
\item \textbf{Selection:} there are typically many variables that one could perturb (e.g. there are roughly 20 000 genes in the human genome), but only a small fraction of them will exhibit surprising interactions. The expert selects pairs of perturbations from this space of all possible pairs.
\end{enumerate}
Measuring interactions is further complicated by the fact that measurement and hypothesis testing are interdependent. We can see this in synthetic lethality example, where if we choose to measure the fraction of cells that survived the respective perturbations, then in order to test independence we would need to test whether $P(\text{survive}_a \cap \text{survive}_b) = P(\text{survive}_a)P(\text{survive}_b)$. 
If we had instead measured the fraction of cells that died, we would have needed to test whether $P(\text{die}_a \cup \text{die}_b) = P(\text{die}_a) + P(\text{die}_b) - P(\text{die}_a)P(\text{die}_b)$. The state of the cells in the respective petri dishes is the same in both cases, but correctly testing (in)dependence depends on how the expert chose to measure that state. These three steps require significant expertise and knowledge to choose and conduct the measurements, which makes scientific discovery hard to automate and scale.

Modern high throughput screening platforms \citep{dove2007high,baillargeon2019scripps, blay2020high, bock2022high, fabio2023high, morris2023next, messner2023mass}  allow us to run large scale perturbation experiments that collect information-rich unstructured measurements. For example, cell painting assays \citep{Bray2016, fay2023rxrx3, Chandrasekaran2023}, provide microscopy images of cells which capture the same cells that the human expert would have measured, but without pre-committing to measuring a particular property of the cell, such as whether it is alive or dead. 
This lets us \emph{measure} all the potential properties of interest at once---%
albeit as unstructured signal data, such as raw pixel images or sensor measurements, rather than preprocessed properties---but it is not clear how to use these measurements for interaction \emph{testing} and \emph{selection} in order to efficiently discover pairwise interactions. 
This suggests the primary question that this paper seeks to address: %
if we dispense with prespecified structured measurements, can we automate the discovery of pairwise interactions by designing testing and selection procedures that efficiently detect interactions from unstructured data?

To test for interactions, 
we show that pairwise perturbations are \emph{separable} if the pairwise experiments provide no additional information beyond what is already known from the single perturbations. 
This separability property can be tested by comparing density ratios of the single and double perturbations to the base / control distribution. 
We then develop an analogous test for the case where pairs of perturbations affect disjoint subsets of the outcome space. For example, if two perturbations affect different organelles in the cell, each of which affect disjoint pixels in our images.  
This strong form of non-interaction is of interest because it gives sufficient conditions for when we can compose summary statistics of the form $\E[h(x)]$ from single perturbations, to predict the corresponding statistics from double perturbations. In particular, this lets us compose embeddings of single perturbations to predict the embeddings of double perturbations in a manner analogous to the classic word-vector analogies \cite{mikolov2013efficient}.

With this notion of an interaction, we address the second task of selection via efficient experimental design. We can search the space of pairwise experiments by selecting pairs of perturbations that are likely to result in large test statistics.
In doing so, we reduce the problem of finding interacting pairs of perturbations into an active matrix completion problem. %
By defining this notion of interaction, we avoid %
the need to characterize uncertainty over the pixel-level outcomes (or some embedding thereof). Instead we directly model the posterior over an unknown reward matrix defined by the test statistics and at every round select actions striking a balance between exploration and exploitation via information directed sampling~\citep[IDS;][]{russo2016information, xu2022adaptive}.

We evaluated our approach on both synthetic and real biological experiments. On the synthetic tasks, we found that our tests were able to detect both forms of dependence, validating our theory. On a benchmark consisting of all pairs of gene knockouts for 50 genes in HUVEC cells, we found that our approach using IDS discovers pairs of genes with higher interaction scores significantly faster than baselines, resulting in $10\%$ more known biological interactions being discovered and potentially far more novel interactions. The interactions we detected were also complementary to those which would have been discovered using existing cosine similarity-based approaches, and as a result, the two approaches can be combined to get a more detailed estimate of the relationships between genes from perturbation experiments.

In summary, we develop a system for discovering pairwise interactions through the following contributions,
\begin{itemize}
    \item We show that two perturbations are separable if and only if, double perturbations provide no additional information than that which was revealed in single perturbations.
    \item We show that two perturbations are disjoint if and only if their density compose additively. This result also implies sufficient conditions for embeddings of perturbations to be composed additively to predict pairwise perturbations.
    \item We show that the test scores for detecting interactions can be used to efficiently search for pairs of perturbations that interact using active matrix completion. 
\end{itemize}

\section{Related work}
\label{sec:related_work}

\paragraph{Causal representation learning}
Our approach to the problem of detecting interactions builds on the modelling assumptions developed in nonlinear independent component analysis \citep{hyvarinen1999nonlinear, hyvarinen2016unsupervised, hyvarinen2019nonlinear} and causal representation learning \citep{scholkopf2021toward}, where we assume we observe some nonlinear mixing function, $g(\cdot)$, of the underlying latent variables. 
The causal representation learning literature typically focuses on disentangling latent variables, whereas we look for testible implications without disentanglement. Our data generating process assumptions are most similar to the interventional setting \citep{ahuja2023interventional, squires2023, buchholz2023learning, varici2024scorebased}.
If we could successfully disentangle latent variables, then our task would be straightforward, but disentanglement is challenging in practice because it is not possible to validate whether an algorithm has succeeded in disentangling latent variables without access to ground truth. Like us, \citet{jiang2023learning} also attempt to learn latent dependence properties (they characterize the whole latent DAG, not just marginal dependence) without disentanglement, but they assume a fixed bipartite graph of dependence between latents and observations, which does not apply to the pixel-level observations that we study. 
\citet{zhang2024identifiability} develop causal representation learning techniques to disentangle latent variables and characterize conditions for extrapolation. Like us, they focus on biological applications, but they rely on stronger polynomial assumptions to achieve disentanglement. Their conditions for extrapolation are dual to our separability tests in that they argue that you can extrapolate when interventions affect \emph{non-overlapping} latents, while we seek to discover when intervention affect \emph{overlapping} latent.
Finally, our separability test is closely related to, and inspired by \citet{wang2023concept}, but whereas they assume separability of concepts to manipulate generative models, we aim to test for an analogous notion of independence in experimental data.

\paragraph{Representation learning for gene knockouts}

Our experiments build on a number of recent works showing the effectiveness of embedding cells presented into a representation space \citet{sypetkowski2023rxrx1, kraus2023masked, Xun2023}. These works show that you can infer that genes code for proteins that form part of complexes by finding embeddings that are highly cosine similar. This works because knocking out proteins on the same pathway will induce the same morphological effect, but this approach is limited to effects that are revealed by single perturbations. 
There have also been a number of papers  that have attempted to learn disentangled representations of cells \citep[e.g.][]{CPA, bereket2023modelling, lopez23a}, mostly from gene expression data.

\paragraph{Design of Gene Knockout Experiments}
There has been considerable work in the recent years to develop algorithms for the design of gene knockout experiments. Typically the problem is studied in the context of discovering single gene-knockouts~\citep{mehrjou2021genedisco} which result in a particular phenotypic effect of interest. A variety of methods including bandits~\citep{pacchiano2023neural} and traditional experimental design~\citep{lyle2023discobax} approaches have been studied in this context. Further, approaches which aim to learn predictors for the effects of unseen gene knockouts typically operate on RNA-seq data~\citep{huang2023sequential}. RNA-Seq data is far more structured than the image data we consider: the former is a high-dimensional vector of count data where each element corresponds to a particular gene's expression, while the latter is just an unstructured collection of pixel intensities.

\paragraph{Experimental design.}
The task of designing of experiments that acquire the most information about the system efficiently %
can be formalized as sequential Bayesian optimal experimental design~\citep[BOED;][]{ryan2016review,foster2021variational,rainforth2023modern}, where the goal to design experiments $x\in\mathcal{X}$ with outcomes $y\in \cal Y$ governed by a generative process $y\sim p(y\mid\gamma, x)$ with parameters $\gamma$. The experiments are performed sequentially $(x_1,\dots, x_T)$, with the objective of maximizing a measure of utility: the information gain~\citep{lindley1956measure,sebastiani2000maximum}. 
While this framework is elegant, it is challenging to apply
when $x$ is high-dimensional because information gain comparisons are very difficult to estimate in high dimensional data. %
We avoid this by defining the explicit task of detecting interactions.

\section{Tests for pairwise interactions}
\label{sec:interaction}

In this section we develop tests for both \emph{separable} and \emph{disjoint} interventions. The
separability test, \cref{sec:separable}, allows us to test whether two
perturbations act on disjoint sets of latent variables. Disjointedness,
\cref{sec:disjoint}, implies compositional generalization of summary
statistics, allowing us to reduce the search space by predicting the
outcome of experiments without explicitly running them.

\paragraph{Setup} We have observations $X$ in an \emph{observation space} $\scX$ of unstructured measurements such as pixels in an image, 
and a finite set of perturbations $\{T_i \in \scT_i: i \in [n]\}$.
Although our theory accommodates generic perturbations, we restrict the discussion to binary perturbations for convenience,
i.e., $\scT_i := \{0, 1\}$ for all $i \in [n]$, and $T_i = 1$ means perturbation $i$ is applied.
For all $i, j \in [n]$, denote the perturbation indicator as follows:
\[
    \delta_0 &= \left\{T_{[n]} = 0\right\}& \delta_i &= \left\{T_i = 1, T_{[n] \setminus \{i\}} = 0\right\} & 
    \delta_{ij} &= \left\{ T_i = T_j = 1, T_{[n]\setminus \{i,j\}} = 0\right\}.
\]
In this section, we assume that for a pair of perturbations $T_i, T_j$, we have access to experimental data
from four distributions: $p(x|\delta_0), p(x|\delta_i), p(x|\delta_j)$, and $p(x|\delta_{ij})$; when we discuss the active learning procedures in Section \ref{sec:active_learning}, we will assume that we have access to all single perturbations distributions, $p(x|\delta_i)$, and we will adaptively select the pairs, $i,j$, on which to collect samples from $p(x|\delta_i, \delta_j)$.
Finally, we assume the existence of a set of latent variables 
$\{Z_1, \dots, Z_L\} \subseteq \scZ$ in some latent space, $\scZ$, that capture
all relevant information about the perturbation. We state this precisely as,
\bassum \label{assump:condindep}
$X \indep (T_1, \dots, T_n) | Z$, or equivalently, 
\[
p(x|T_1, \dots, T_n) = \int_{\scZ} p(x|z) p(z| T_1, \dots, T_n) \d z.
\]
\eassum
Throughout, we assume all distributions have well-defined densities or probability mass functions with respect to some $\sigma$-finite base measure on their corresponding sample spaces.
We use the same symbol to denote a distribution and its density. 
All proofs are deferred to \cref{apdx:proof}.

\subsection{Separability testing} \label{sec:separable}

It would be trivial to verify whether two perturbations intervene on different latents
if the underlying latent variables were observed directly. 
However, our observations are unstructured signal data, which we assume is generated by some deterministic mixing function.
\bassum \label{assump:diffeo}
There exists a diffeomorphism\footnote{A diffeomorphism is a differentiable
bijection with a differentiable inverse.} $g: \scZ \to \scX$ such that $X = g(Z)$. 
\eassum
\brmk
This diffeomorphism assumption between $\scZ$ and $\scX$ can be relaxed; 
our theory and methodology remain valid as long as the change of variable formula holds for the distributions of $Z$ and $X$. 
For example, the distribution of $X$ may have support on a low-dimensional manifold of $\scX$, with the latent space $\scZ$ having a lower dimension than $\scX$.
See \citet[Lemma 5.1.4]{krantz2008geometric} for the generalized change of variable formula for non-bijective transformations.
\ermk

The key observation that we leverage is that when latent variables are independent, the change of variable formula implies that the density ratio of a
perturbed distribution to the original (control) distribution has a simple form that only involves the distribution of the intervened \emph{latent} variable,
\[\label{eq:densityratio}
\frac{p(x | \delta_i)}{p(x|\delta_0)} = \frac{p_Z(g^{-1}(x)| \delta_i)\left | \text{det}(J(g^{-1}(x))) \right|}{p_Z(g^{-1}(x)|\delta_0)\left | \text{det}(J(g^{-1}(x))) \right|} =  \frac{p^\dagger_{Z_i}(g^{-1}(x))}{p_{Z_i}(g^{-1}(x))}.
\]
Here $p^\dagger_{Z_i}$ denotes the perturbed distribution of the latent
variable $Z_i$ targeted by intervention $i$. The testable conclusion that we can
derive from this observation, is that when two variables are independent, we can
predict the density ratio of the double perturbation, 
$\frac{p(x | \delta_{ij})}{p(x|\delta_0)}$, as the product of the density ratios of the respective single perturbations, 
$\frac{p(x | \delta_{i})}{p(x|\delta_0)}\frac{p(x | \delta_{j})}{p(x|\delta_0)}$. 
To make this rigorous, we assume that there exists a causal factorization of the
latent distribution, where each latent variable is conditionally independent of
its non-descendents given its parents. We use $\pa(Y)$ (resp. $\ch(Y)$) to denote the parent (resp. children) of 
a random variable $Y$.

\bassum \label{assump:causaldag}
The latent distribution $p_Z(z)$ can be factorized into $L$ latent factors:
\[
    p_Z(z| T_1, \dots, T_n) = \prod_{l = 1}^L p_{Z_l}(z_l|\pa(z_l)), 
\]
where $\forall l \in [L]$, $\pa(Z_l) \subseteq T_{[n]} \cup Z_{[L]}$, 
and $\forall i \in [n]$, $\ch(T_i) \subseteq Z_{[L]}$.
The random variable $Z$ here should be interpreted as the concatenation of all latent factors $\{Z_1, \dots, Z_L\}$.
\eassum

The observed effects of two non-interacting perturbations are attributed to distinct causal pathways, 
rather than being confounded by a shared latent factor.
\bdefn
Two perturbations $\delta_i, \delta_j$ are \textbf{separable} if $\mathrm{Ch}(T_i) \cap \mathrm{Ch}(T_j) = \emptyset$. 
\edefn

When two perturbations act separably, the resulting density ratios of the observation distribution will have the predictable interactions that we derived above.
\bthm
\label{thm:lpdfscoreadd}
Suppose that \cref{assump:causaldag,assump:diffeo} hold, and that $\delta_i, \delta_j$ are separable. Then, 
\[\label{eq:separabldensityratio}
    \frac{p(x | \delta_{i,j})}{p(x|\delta_0)} = \frac{p(x | \delta_{i})}{p(x|\delta_0)}\frac{p(x | \delta_{j})}{p(x|\delta_0)}.
\]
\ethm

This provides a testable implication of separability, but to test it, we need to derive a real-valued test statistics. By taking logs and rearranging terms, we can rewrite \cref{eq:separabldensityratio}, as testing, 
\[
\label{eqn:log_rel}
    \log p(x|\delta_{ij}) + \log p(x|\delta_0) -  \log p(x|\delta_{i}) - \log p(x|\delta_{j}) \stackrel{?}{=} 0
\]
If we instead take expectations of \cref{eqn:log_rel} with respect to $p(x|\delta_0)$, then this
amounts to testing if 
\[ \label{eq:kltest}
\kl{ p_0}{p_{i}} + \kl{ p_0}{p_{j}} \stackrel{?}{=} \kl{ p_0}{p_{i,j}},
\]
where $p_{i} = p(x|\delta_i), p_0= p(x|\delta_0)$ and $\kl{\cdot}{\cdot}$ denotes the Kullback–Leibler (KL) divergence.
Note that KL divergence measures the difference between two distributions; in our context, it quantifies the distribution shift resulting from interventions. 
\cref{eq:kltest} reflects the intuition that the influence of two separable perturbations, measured by KL divergence, is additive.
We use the KL score, $\left| \kl{ p}{p_{i}} + \kl{ p}{p_{j}} - \kl{ p}{p_{i,j}}\right|$, to quantify the violation of the separability of $\delta_i, \delta_j$, 
where these KL divergences are estimated using samples. 

In practice, we observe that the simple K-nearest neighbor-based (KNN) estimator of
the KL divergence \citep{wang2009divergence} suffices for
low-dimensional problems. However, for high-dimensional data such as images, more dedicated estimation procedures are required \citep{belghazi2018mutual,songunderstanding,ghimire2021reliable}. 
We detail our estimation procedure in \cref{apdx:klest}.

\subsection{Disjointedness testing} \label{sec:disjoint}

Our second testing procedure examines whether two perturbations operate on disjoint domains such that their effects are cumulative. 
For example, if the morphological changes from knocking out two non-interacting genes can be separated into distinct visual features, such that their measures sum, then they are disjoint. %
We can define disjointedness of two perturbations formally as,

\bdefn \label{def:disjoint}
Two perturbations $\delta_i, \delta_j$ are \textbf{disjoint} if 
\[ \label{eq:disjoint}
    p(x|\delta_{ij}) - p(x|\delta_0) = (p(x|\delta_{i}) - p(x|\delta_0)) +  (p(x|\delta_{j}) - p(x|\delta_0)).
\]
\edefn

Disjointedness is important, because it implies that we can predict pairwise
summary statistics of the distributions from individual perturbations. To see
this, let $h(\cdot)$ denote \emph{any} feature map from the observation, $X$.
If we have disjoint perturbations $\delta_i, \delta_j$, then,
\[
    \E[h(x) | \delta_{i,j}] - \E[h(x) | \delta_{0}] &= \int h(x) [p(x|\delta_{ij}) - p(x|\delta_{0}) ] \d x\\
    &= \int h(x) [(p(x|\delta_{i}) - p(x|\delta_0)) +  (p(x|\delta_{j}) - p(x|\delta_0))] \d x\\
    & = \E[h(x) | \delta_{i}] - \E[h(x) | \delta_{0}] + \E[h(x) | \delta_{j}] - \E[h(x) | \delta_{0}]. \label{eq:embeddingtest}
\]

This implies that we can define average centered embedding vectors, $\vec{h}_i :=  \E[h(x) | \delta_{i}] - \E[h(x) | \delta_{0}]$ and $\vec{h}_j$, 
and accurately predict $\vec{h}_{i,j} = \vec{h}_{i} + \vec{h}_{j}$ without running the experiments. 
It is worth noting that there is growing evidence in the literature \citep{lotfollahi2019scgen, gaudelet2024season} that shows that these relationships often hold in real biological experiments (and our experiments support this). 
Disjointedness explains the sufficient conditions for this to hold. \cref{apdx:chooseh} explains the choice of $h$ we tested.

We can test whether \cref{eq:disjoint} holds, by testing the following null hypothesis:
\[
    \mathrm{H}_0 : \frac{1}{2} p(x|\delta_{ij}) + \frac{1}{2} p(x|\delta_0)  = \frac{1}{2} p(x|\delta_{i}) + \frac{1}{2} p(x|\delta_{j}).
\]
Given interventional data from $p(x|\delta_0), p(x|\delta_i), p(x|\delta_j)$ and $p(x|\delta_{ij})$, 
we can frame this as a standard two-sample test problem. 
With balanced experiments, i.e., $p(\delta_0) = p(\delta_i) = p(\delta_j) = p(\delta_{ij})$, 
we can create samples from the mixture $\frac{1}{2} p(x|\delta_{ij}) +
\frac{1}{2} p(x|\delta_0)$ by combining data from the controlled and doubly perturbed groups. 
Similarly, we can obtain samples from$\frac{1}{2} p(x|\delta_{i}) + \frac{1}{2} p(x|\delta_{j})$. 
When experiments are heavily unbalanced, we can balance the datasets through downsampling or upsampling.

In practice, we employ the maximal mean discrepancy (MMD) based two-sample test \citep{gretton2012kernel}, 
which compares two distributions based on their embeddings in some reproducing kernel Hilbert space (RKHS).  
The estimated MMD serves as a measure of the extent to which the perturbations violate the disjointedness. 
Interestingly, MMD test amounts to test \cref{eq:embeddingtest} on a ``most discriminative'' feature map $h$ in the RKHS.
We provide a self-contained introduction about MMD and kernel mean embedding of
distributions in \cref{apdx:mmd}.

Finally, it is helpful to consider a concrete generative process that yields disjoint perturbations. 
Unlike the separability formulation, which assumes the existence of the Markov factorization of the latent distribution 
(Assumption \ref{assump:causaldag}), the disjointedness models the latent distribution as a finite mixture. 
In this framework, non-interacting perturbations intervene on different mixing components.

\bassum \label{assump:mixture}
The latent distribution $p_Z(z)$ admits the form of an $L$-component mixture:
\[
    p_Z(z| T_1, \dots, T_n) = \sum_{l = 1}^L w_{l} \cdot p_{Z_l}(z|\pa(z_l)), \quad \text{where } (w_1, \dots, w_L) \in \Delta_{L}.
\]
In addition, perturbations do not intervene the mixing weights $(w_1, \dots, w_L)$. 
\eassum
\bthm \label{thm:mixture}
Suppose that \cref{assump:condindep,assump:mixture} hold. Then $\delta_i, \delta_j$ are disjoint
if $\ch(T_i) \cap \ch(T_j) = \emptyset$.
\ethm

\section{Selecting perturbation pairs to efficiently discover interactions}
\label{sec:active_learning}
With principled frameworks for testing separability and disjointedness in place, the natural next question is \emph{how to select the experiments to run}? The space of possible pairs of perturbations is typically too large to perform experiments on all pairs as each experiment is costly. For example, pairwise knockouts of all 20 000 genes on the human genome would require approximately 200 million experiments (not including replicates). In this section we discuss how tools from experimental design and bandits can be used to efficiently select pairs of perturbations likely to have interactions. 

\paragraph{Selection of perturbation pairs as active matrix completion.}

The testing frameworks discussed in Section~\ref{sec:interaction} prescribe test statistics which can be used to detect pairwise interactions. The test statistics, however, require samples from $p(x|\delta_{i,j})$, which entails running pairwise perturbation experiments. We are thus interested in developing an approach for selecting pairs of perturbations which are \emph{likely} to have high values for the test statistics and are thus likely to reveal pairwise interactions. 

The test statistics for pairs of perturbations can be viewed as an (unknown) symmetric matrix $\mathbf{R} \in \mathbb{R}^{n\times n}$, where each entry $\mathbf{R}_{i,j}$ contains the value of the test statistic that we will observe if we run perturbation $\delta_{ij}$. $\mathbf{R}$ is unknown a priori but we are allowed to select entries to observe. This can be viewed as an \emph{active matrix completion} problem~\citep{chakraborty2013active}. Active matrix completion is a variant of the standard matrix completion problem~\citep{laurent2009matrix} where values of entries of the matrix can be sequentially queried. Framing the problem of selecting perturbation pairs as an active matrix completion allows us to leverage existing efficient algorithms. 

\begin{algorithm}[t]
  \caption{ASD for selecting perturbation pairs}
  \label{alg:discovering_pairs}
\begin{algorithmic}[1]
\STATE Initialize $H_1 = \{\}$, batch size $b$
\FOR{$t=1,\dots, T$}
\STATE Estimate posterior $p(\mathbf{R}\mid H_t)$
\STATE Compute information ratio $\Psi_t$
\STATE Pick batch $(a_1, \dots, a_b)$ greedily which minimize $\Psi_t$
\STATE Perform experiments and compute $\{\mathbf{R}_{a_1},\dots, \mathbf{R}_{a_b}\}$
\STATE Update $H_{t+1} = H_t \cup \{(a_1, \mathbf{R}_{a_1}, \dots, (a_b, \mathbf{R}_{a_1})\} $
\ENDFOR
\end{algorithmic}
\end{algorithm}

\paragraph{Adaptive sampling for discovery.} In particular, we use the framework of \emph{adaptive sampling for discovery}~\citep[ASD;][]{xu2022adaptive} which provides a bandit-based approach for active matrix completion. ASD involves using \emph{information directed sampling}~\citep[IDS;][]{russo2016information} in a ``discovery'' setting, where an action is only selected once. This is an instantiation of the general sleeping experts setting~\citep{kanade2009sleeping}, where the set of available actions shrinks every round. %

Let $\Delta$ denote the action space of possible experiment designs, which in this case is the set of perturbations $\Delta := \{\pert_{ij}: i, j\in n, i > j\}$, where $n$ is the total number of distinct perturbations.To simplify notation, we denote the perturbation pair $i,j$ selected at step $k$ as $a^{(k)}:= (i^{(k)},j^{(k)}) \in \Delta$. $\mathcal{D}(\Delta)$ is the set of possible (categorical) distributions defined over $\Delta$. Each time an action, $a^{(k)}$, is selected, the corresponding element of the (unknown) reward matrix, $\mathbf{R}$, is revealed to the agent. In our setting this reward matrix corresponds to the test statistic from either Section \ref{sec:separable} or \ref{sec:disjoint} for each pair of perturbations. Let $H_{t} = ((a^{(k)}, \mathbf{R}_{i^{(k)}, j^{(k)}}))_{k=1}^{t-1}$ denote the history of actions and their corresponding rewards until round $t$, and $\Delta_{t}$ denotes the set of remaining actions at round $t$; i.e. the pairs of perturbations that we have not yet tested experimentally. A policy $\pi$ is defined as a map from $H_t$ to $\mathcal{D}(\Delta_t)$. The IDS policy, $\pi_\text{IDS}$ maintains a posterior distribution over $\mathbf{R}$ given the data observed up to round $t$, which we denote $p(\mathbf{R}\mid H_{t})$. 

We can describe the sub-optimality of any action with respect to a set of beliefs by comparing the action's reward to that of the best action that could have been selected at time $t$, under the agent's current posterior over the reward matrix. Intuitively, we can evaluate this by sampling a plausible reward matrix from our posterior, $\dot{\mathbf{R}} \sim (\mathbf{R}|H_{t})$, and then comparing the reward from action, $a$, to the reward an agent would have received from selecting the optimal action, $a^* = \argmax_{a\in \Delta_t} \dot{\mathbf{R}}(a)$; where $\dot{\mathbf{R}}(a^{(k)}):= \dot{\mathbf{R}}_{i^{(k)}, j^{(k)}}$ for $a^{(k)} = (i^{(k)}, j^{(k)})$. This is known as the expected instantaneous regret incurred by an action and is defined as, 
\[
\Delta_t(a) = \mathbb{E}_{\dot{\mathbf{R}}\sim p(\mathbf{R}|H_{t})}[ \dot{\mathbf{R}}(a^*) - \dot{\mathbf{R}}(a)].
\]
Additionally, we define the information gain about the top $T-t+1$ remaining actions as follows: 
\[
g(a) = MI(a_{t,1}^*, \dots, a_{t, T-t+1}^*;\mathbf{R}(a)\mid H_t, \delta_t=a).
\]

Algorithm~\ref{alg:discovering_pairs} summarizes the algorithmic procedure. The algorithm operates over a series of $T$ rounds. In each round, the first step is to estimate the posterior distribution over $\mathbf{R}$ given the data observed thus far. Next step involves selecting a batch of perturbations based on the information ratio. The IDS policy at round $t$ can then be computed by minimizing the \emph{information ratio} $\Psi$:
\[
    \pi_\text{IDS} \in \arg\min_{\pi \in {\cal D}(\Delta_t)} \Psi_{\pi, t} \coloneqq \frac{(\Delta_t^\top\pi)^\lambda}{g_t^\top\pi}
\]
where $\lambda$ is a parameter controls the tradeoff between lower instant regret (exploitation) and higher information gain (exploration). As $g_t(a)$ is intractable to compute in general, following~\citet{russo2016information,xu2022adaptive} we use an approximation, replacing $g_t$ with the conditional variance $v_t(a) = \text{Var}_t(\mathbb{E}[\mathbf{R}_a|a_1^*, a])$. The conditional variance is a lower bound on the information gain $g_t(a) \ge v_t(a)$ and can thus replace the mutual information (since we are interested in maximizing it). After selecting the perturbation pair $i, j$, we compute the pairwise test statistic, $\mathbf{R}_{i,j}$ using data from the experimental outcomes. We add these test statistics to our reward matrix, update our posteriors and then continue on to the next round.  

Without any structural assumptions on $\mathbf{R}$, linear regret is unavoidable for any bandit algorithm. We thus make an assumption that $\mathbf{R}$ is low-rank with a Gaussian prior on the columns. Under this assumption, ASD achieves sublinear regret~\citep{xu2022adaptive}.

\paragraph{Batching.}
In high-throughput experimental screens, it is possible to run multiple experiments in parallel at the cost of a single experiment. So instead of a single action $\delta_t$, we can select a set of actions $\{\delta_t^1, \dots, \delta_t^b\}$ where $b$ is the number of experiments we can run in parallel. However, as there are no efficient algorithms for combinatorial bandits  in the discovery setting, we resort to a simple greedy scheme to select batches with ASD. Specifically, instead of picking a single action which minimizes the information ratio, we pick $b$ actions with the lowest information ratio.

\section{Experiments}
\label{sec:experiments}

Our experiments aim to address three objectives: (1) verifying that our theoretical claims about
interactions are detectable on known synthetic tasks; (2) evaluating the test
statistics' ability to recover known biological relationships from real
pairwise perturbation experiments; and (3) assessing our active learning
pipeline's efficiency in detecting interactions.

In all the experiments, 
our MMD-based tests used the RBF and Matern 2.5
kernels, and we chose bandwidth using median heuristics. 
Unless otherwise stated, we estimated the KL score by first learning the three log-density ratios 
$\log \frac{p(x|\delta_{ij})}{p(x|\delta_0)}, \log \frac{p(x|\delta_{i})}{p(x|\delta_0)}, \log\frac{p(x|\delta_{j})}{p(x|\delta_0)}$ using contrastive learning \citep[NRE]{hermans2020likelihood},
and then obtaining the KL estimates via 
the smoothed mutual information ``lower-bound'' estimator (SMILE) \citep{songunderstanding} with clipping parameter $\tau = 5$. 
Detailed explanation about the NRE log-density ratio estimator and SMILE are provided in \cref{apdx:klest}.
Additional experimental details are provided in \cref{apdx:expt}.

\subsection{Testing on synthetic setting}

We validated the separability test on two synthetic interventional examples:
one with 3-dimensional tabular data and one with images (of size $3\times 128 \times 128$).
We also validated the disjointedness test on an interventional tabular example.
In each example, we generated observations by sampling from a latent distribution $p_Z$ that obeys a DAG structure,  
followed by a mapping $g(\cdot)$ that maps the latent samples to the observations.
We then estimated the test statistics for the corresponding test for each pair of perturbations.
Detailed descriptions of the data generating processes for the three examples, the DAG structures of the latent distributions
$p_Z$ and the mapping from latents to observations $g(\cdot)$, 
are provided in \cref{apdx:sync}.

\begin{figure}
    \centering
    \includegraphics[width=0.325\textwidth,trim={2cm 2cm 2cm 2cm},clip]{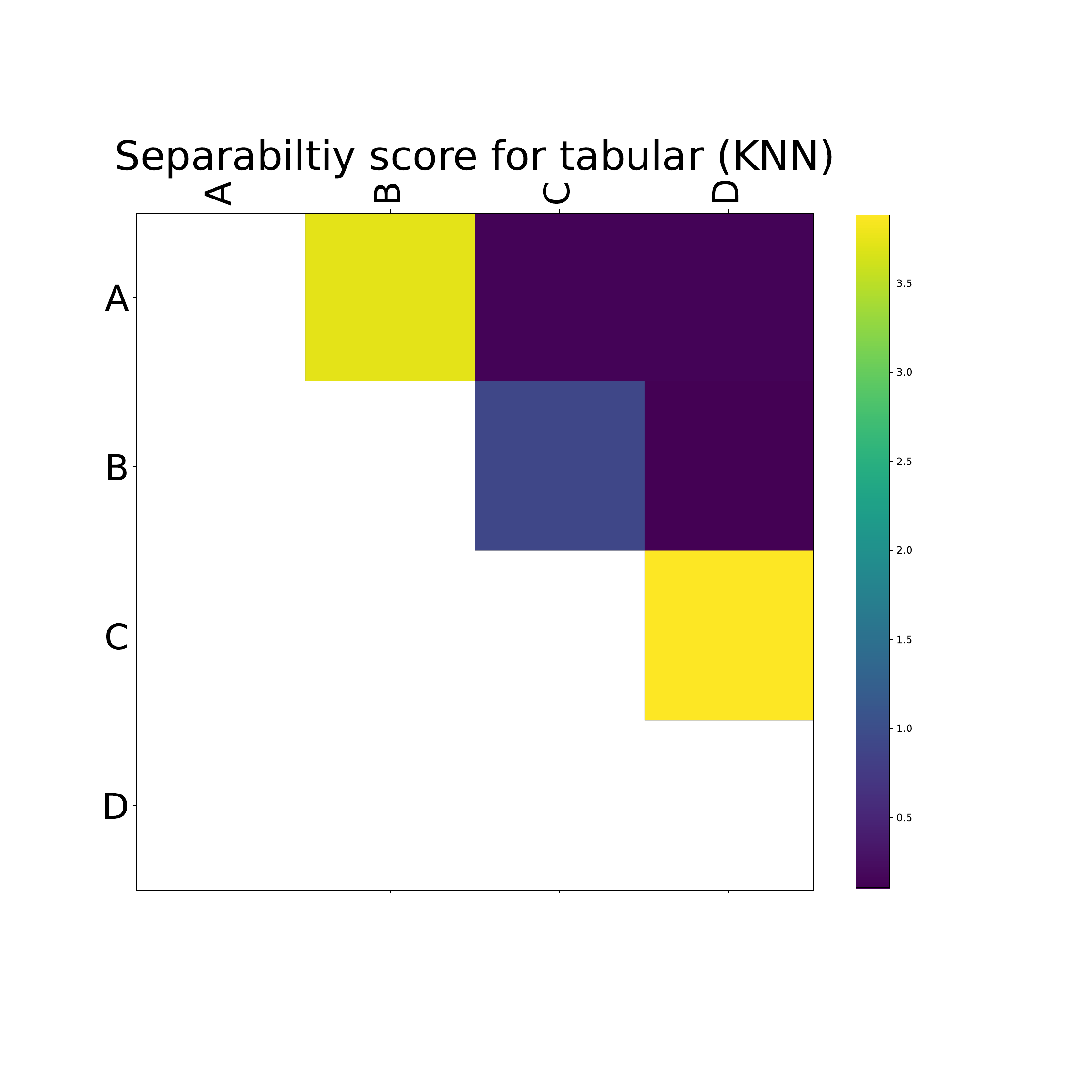}
    \includegraphics[width=0.325\textwidth,trim={2cm 2cm 2cm 2cm},clip]{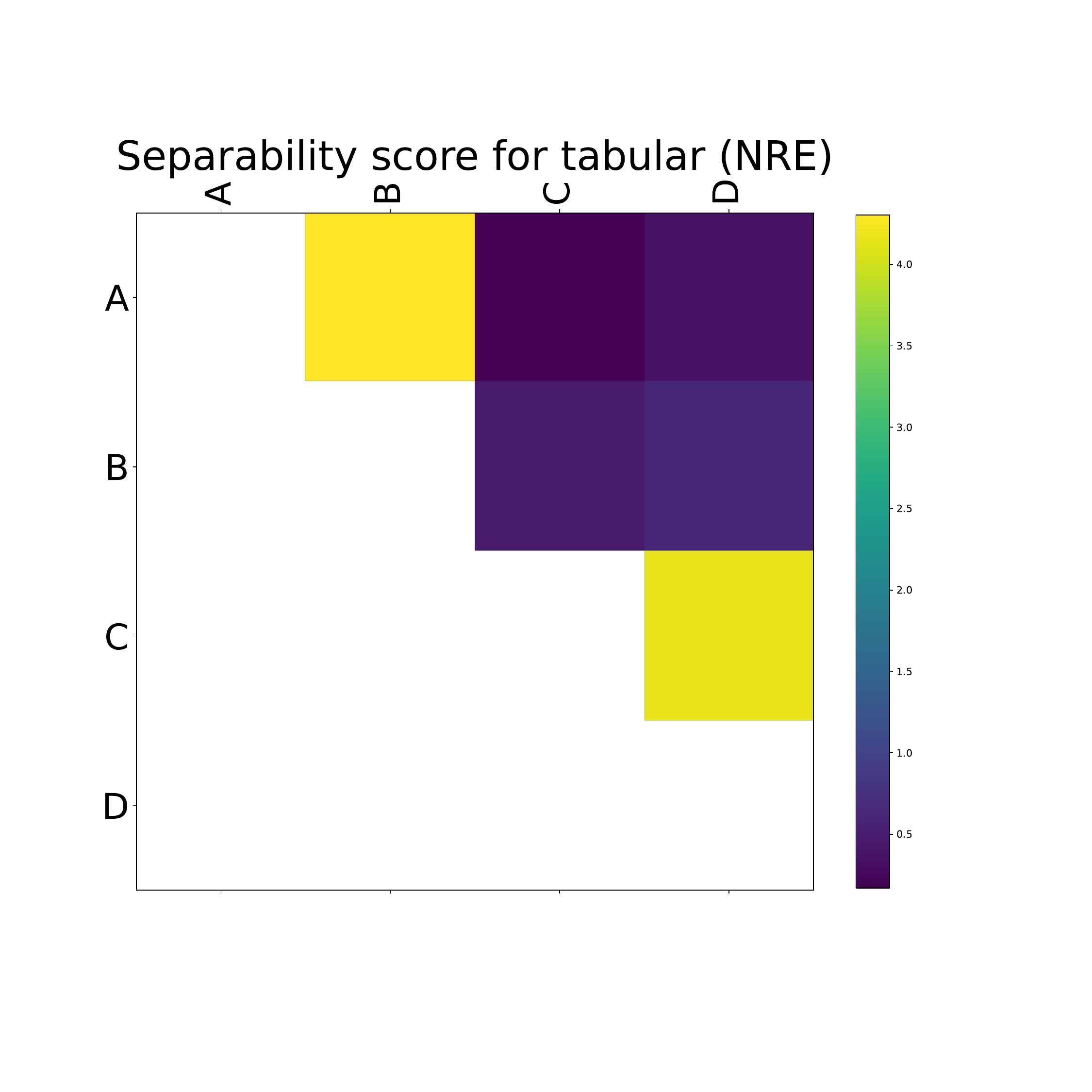}
    \includegraphics[width=0.325\textwidth,trim={2cm 2cm 2cm 2cm},clip]{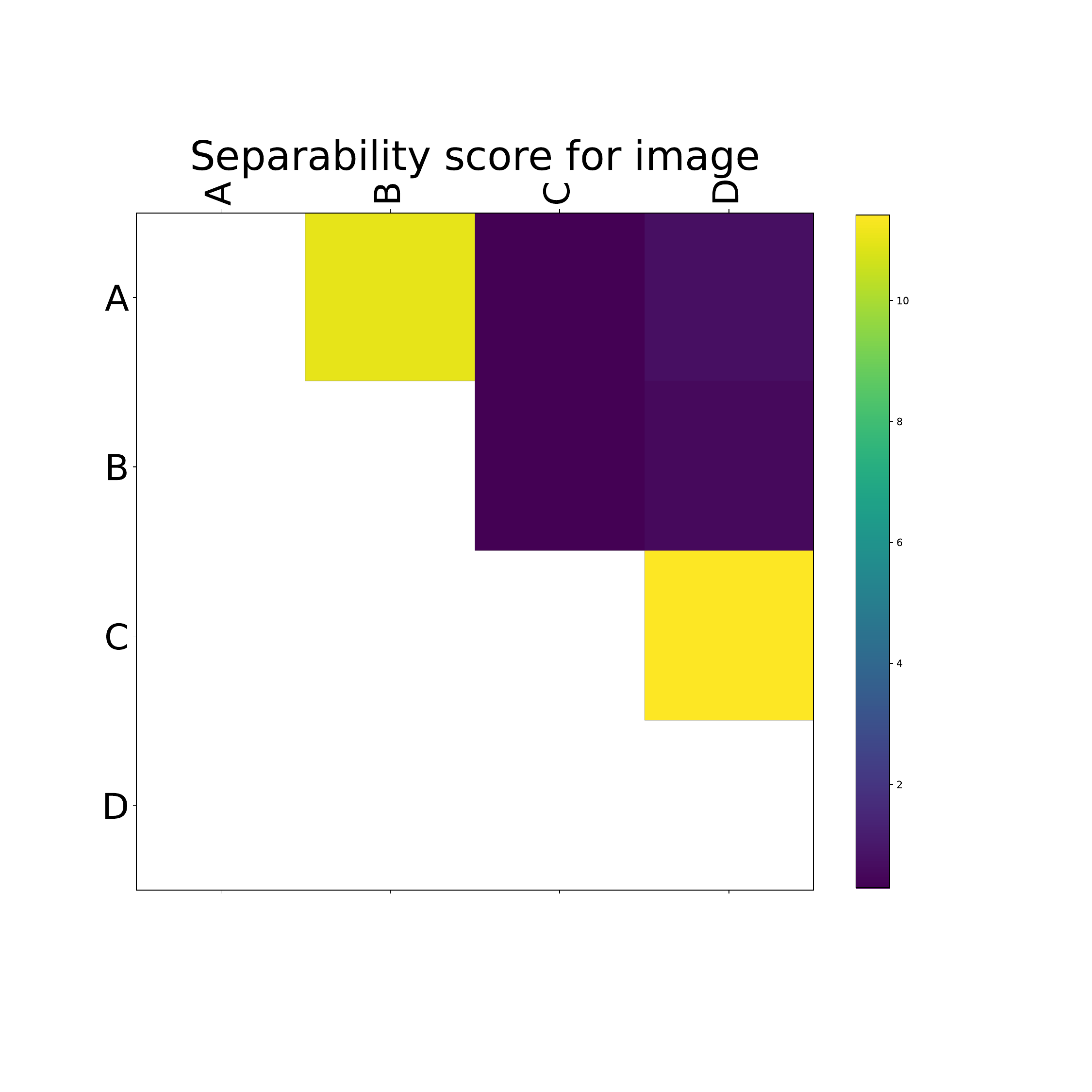}
    \caption{
    Separability testing on both the synthetic tabular data using KNN-based KL estimator (\emph{left}) and NRE-based KL estimator (\emph{middle}), and the synthetic images (\emph{right}); brighter colors suggest stronger interactions.
    Ground truth interacting pairs for both examples are A-B and C-D, which are correctly identified.
    }
    \label{fig:sepsync}
\end{figure}

\begin{figure}
    \centering
    \includegraphics[width=0.45\textwidth,trim={3cm 3cm 3cm 4cm},clip]{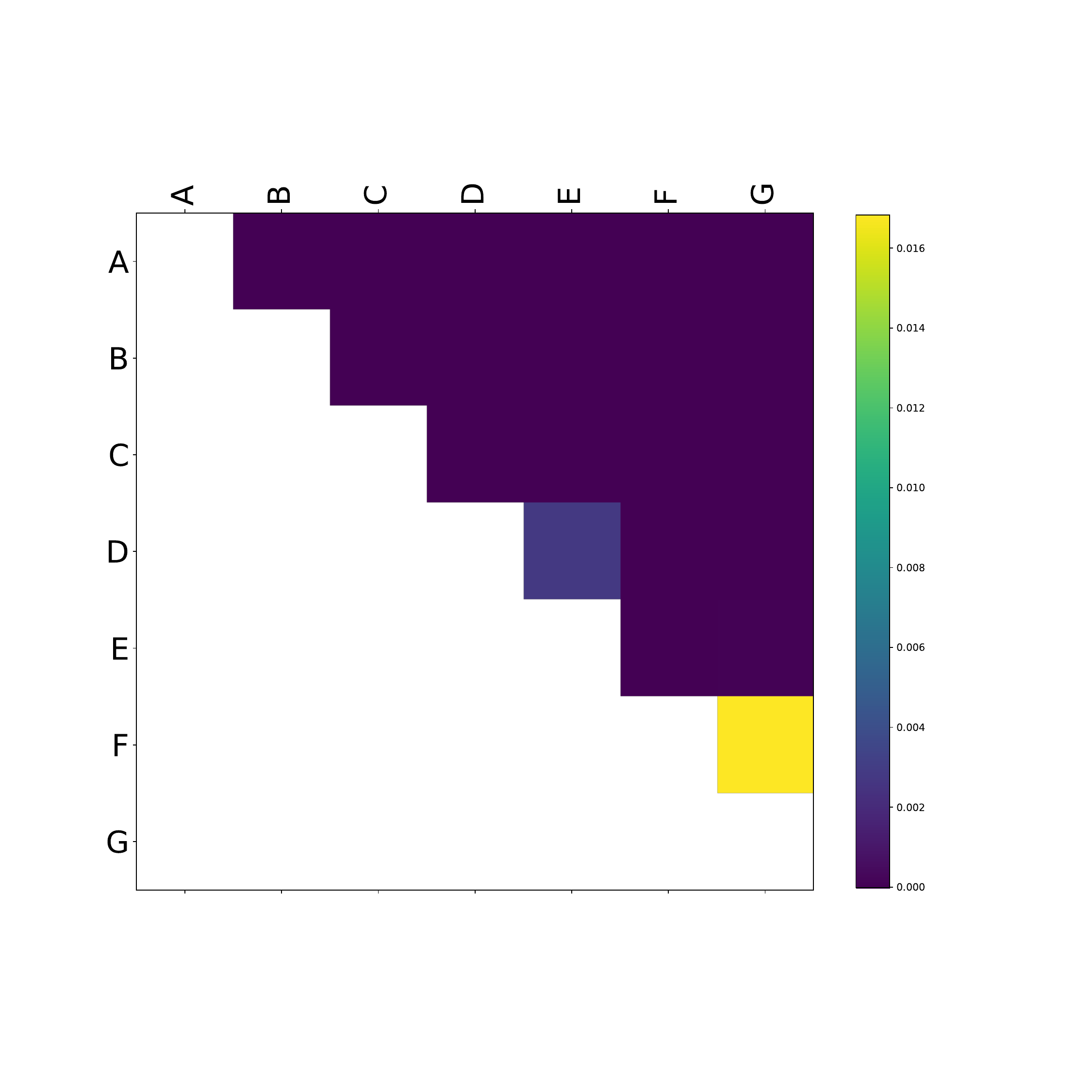}
    \includegraphics[width=0.45\textwidth,trim={3cm 3cm 3cm 4cm},clip]{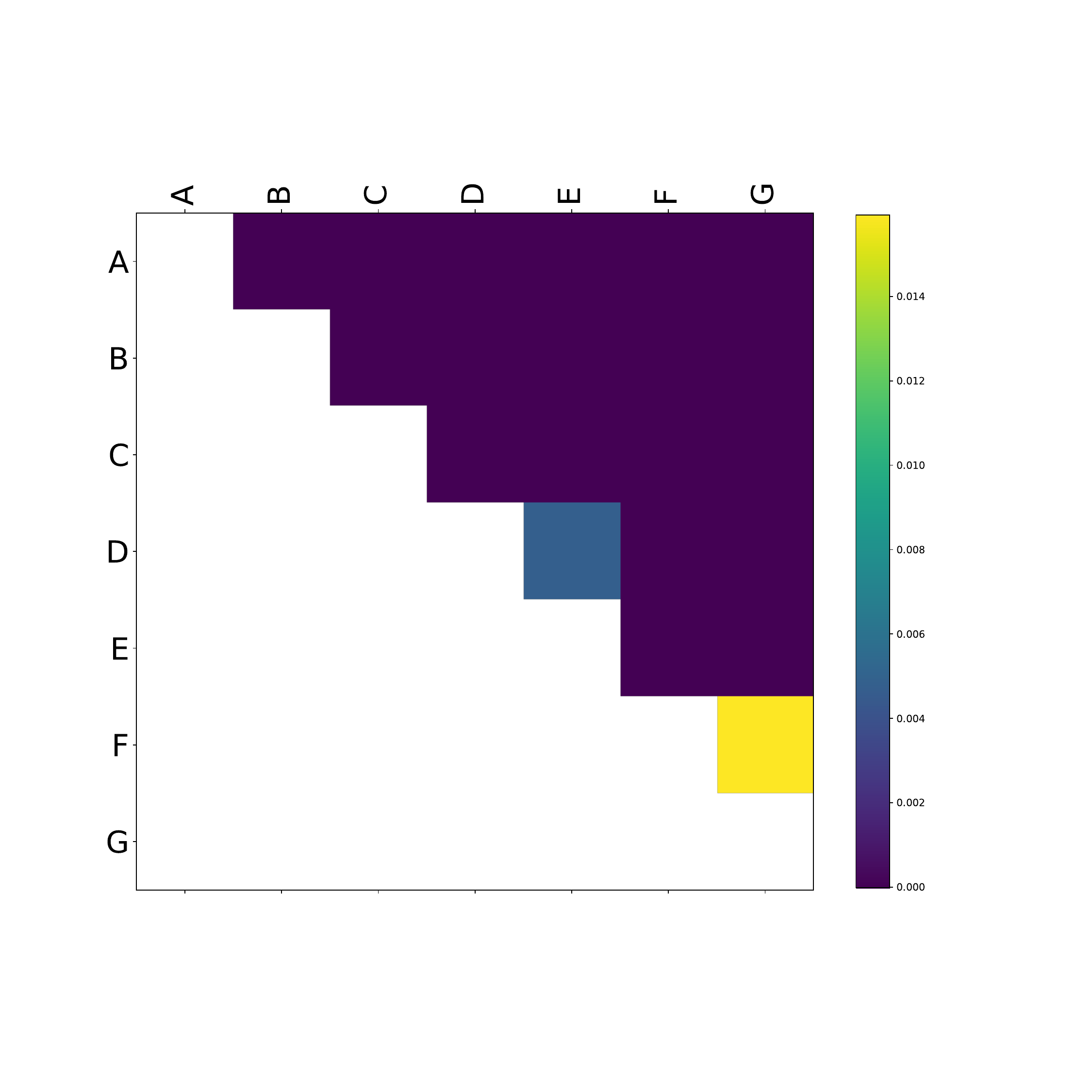}
    \caption{
    Disjointedness testing on synthetic example using the MMD-based statistics with a Matern 2.5 kernel (\emph{left}) and an RBF kernel (\emph{right}); brighter colors suggest stronger interactions.
    The ground truth interacting pairs are D-E and F-G, which are correctly identified by the test.
}
    \label{fig:dissync}
\end{figure}

\cref{fig:sepsync} shows the synthetic results for the separability test on both
the tabular and image data. The results indicate that, in both examples, our estimated KL score
accurately characterized the separability relationships between perturbations:
inseparable pairs result in large KL scores, while separable
pairs result in small scores. The tests for disjointedness are shown in \cref{fig:dissync},
demonstrating that the MMD test accurately identified failures of
disjointedness and is relatively insensitive to the choice of kernel.

\subsection{Testing on Biological interactions}

\begin{figure*}[h]
    \centering
    \includegraphics[width=0.49\textwidth]{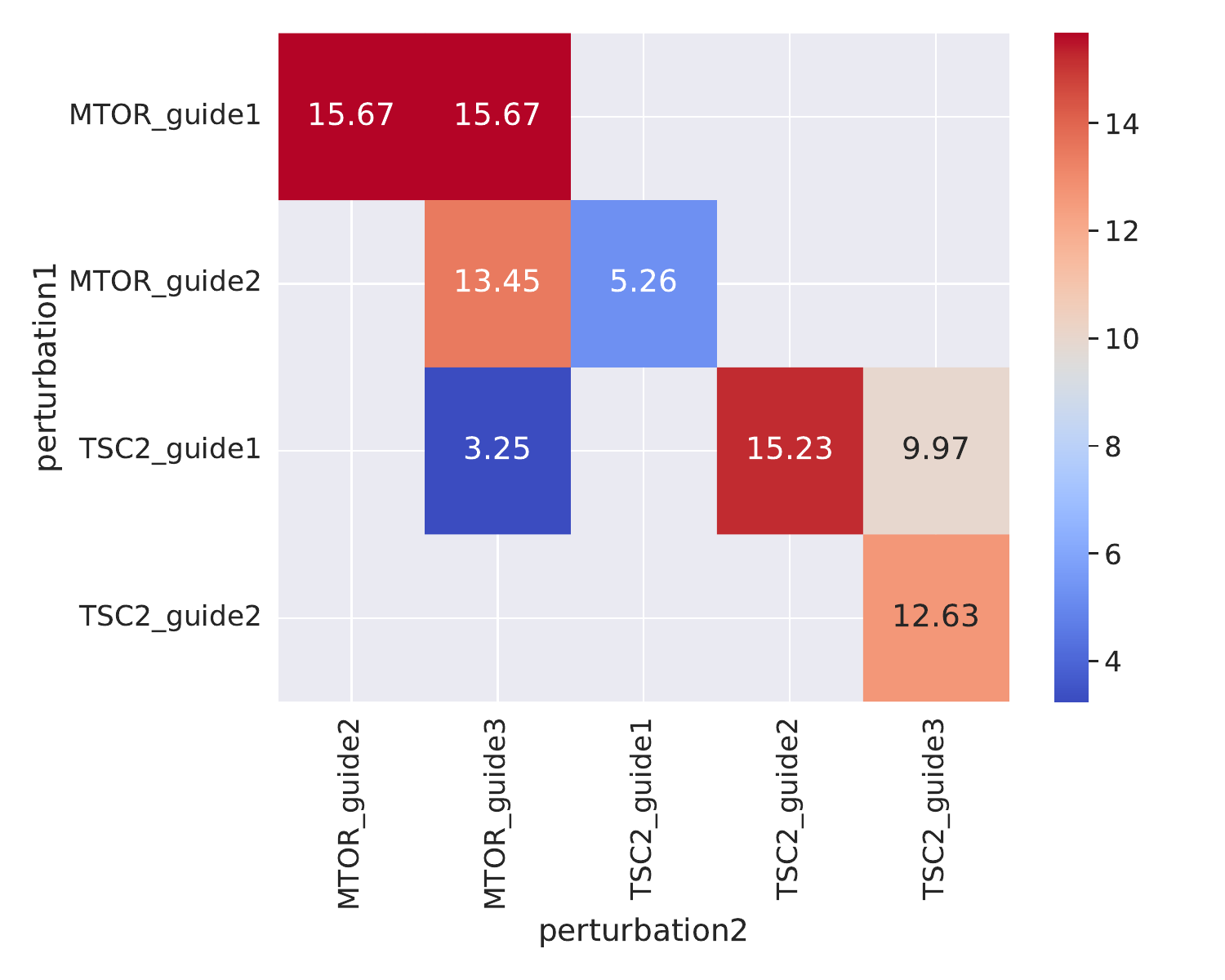}
    \includegraphics[width=0.49\textwidth]{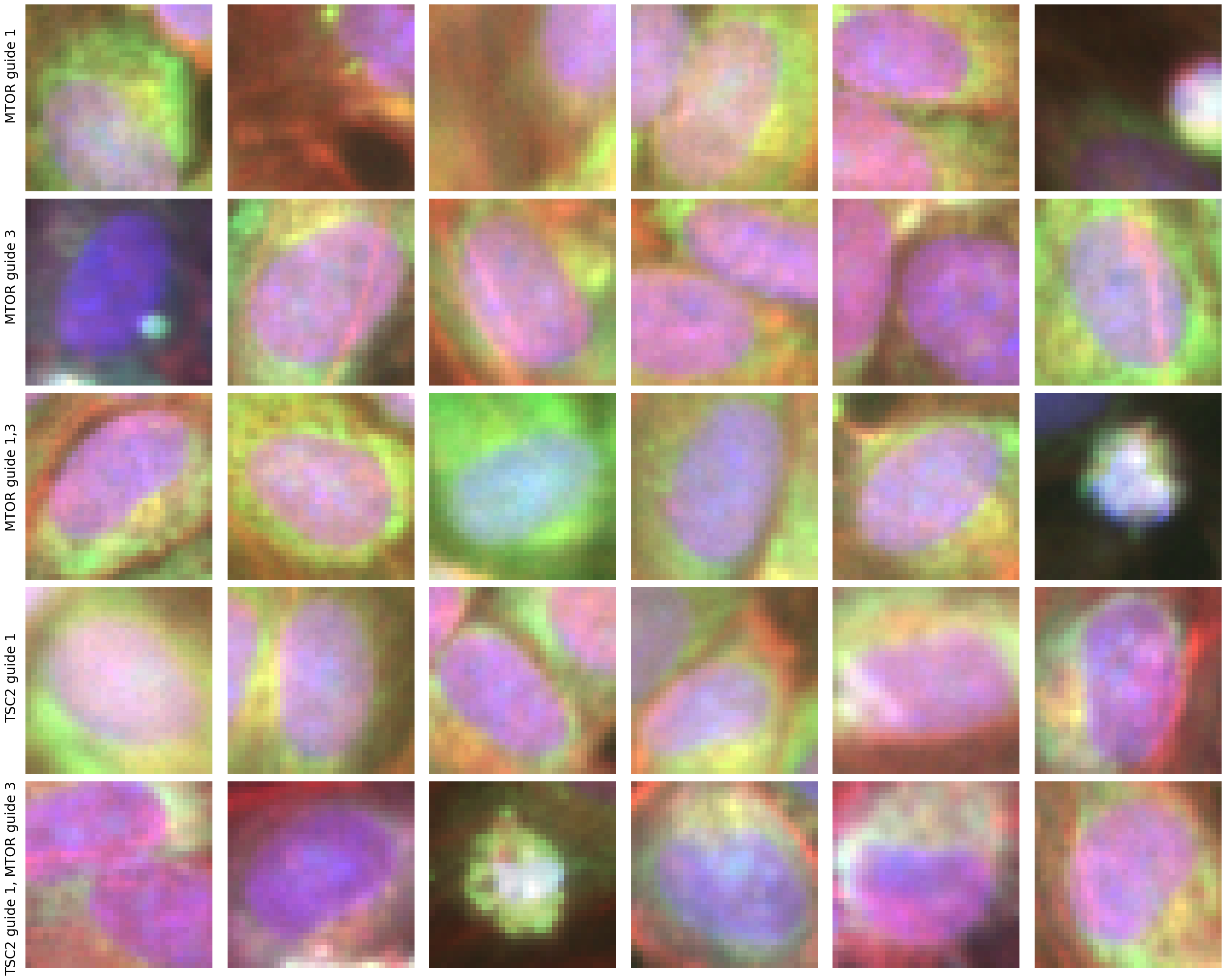}
    \caption{
        \emph{(left)} Pairwise separability scores between different CRISPR guides of two genes, TSC2 and MTOR. 
        Missing pairs means that the data for corresponding pairwise combination was not collected in our biological experiment.
        Guides targeting the same gene show high KL scores (red), while guides targeting different genes show low scores (blue).
        \emph{(right)} Random samples of the actual single cell images used in these experiments. Note that detecting the presence or absence of an interaction is extremely difficult, even for trained experts. 
    }
    \label{fig:pp}
\end{figure*}

In order to evaluate whether our separability test can recover known biological interactions, we first ran the following test. 
In gene knockout experiments, one can target the same gene with multiple different CRISPR guides, each of which cuts the gene in different places, but (at least in theory) results in the same gene being knocked out. 
Intuitively, guides targeting the same gene should show high scores in the separability test because they are targeting the same latent variable, while guides targeting different genes should show lower scores (assuming the genes are on distinct pathways). 
While two different guides that target the same gene are not usually run in a single well, we have in our dataset a couple of examples of this data for two genes (TSC2 and MTOR). For this experiment, we used single-cell cell painting images and tested the separability between pairs of guides.
The matrix of the separability scores displayed in \cref{fig:pp} was consistent with what we would expect: we observed strong interaction scores between guides targeting the same gene (e.g. MTOR guide 3 with both MTOR guides 1 and 2), and much weaker scores for the interaction between the MTOR and TSC2 targeting guides. It is worth noting that both MTOR and TSC2 affect many systems within the cell, so we should expect some interaction, but the fact that the interaction was far smaller than the interaction score for guides targeting the same gene is very encouraging.

We then evaluated our testing approaches on a collection of 50 genes with the
goal of detecting gene interactions. The dataset was collected by performing
CRISPR knockouts on all pairs from a set of 50 selected genes, resulting in
1,225 gene pairs. The targeted genes have a bias towards known
gene-gene interactions.
We performed \textit{in vitro} double gene knockout experiments on HUVEC cells using
three CRISPR guides per gene. We label perturbations according to the targeted gene, aggregating the effects of the individual CRISPR guides. The experimental protocol and data preprocessing followed the procedures
described by \citet{fay2023rxrx3} and \citet{sypetkowski2023rxrx1},
respectively.

\begin{figure*}[h]
    \centering
    \includegraphics[width=0.325\textwidth,trim={4cm 2cm 4cm 4cm},clip]{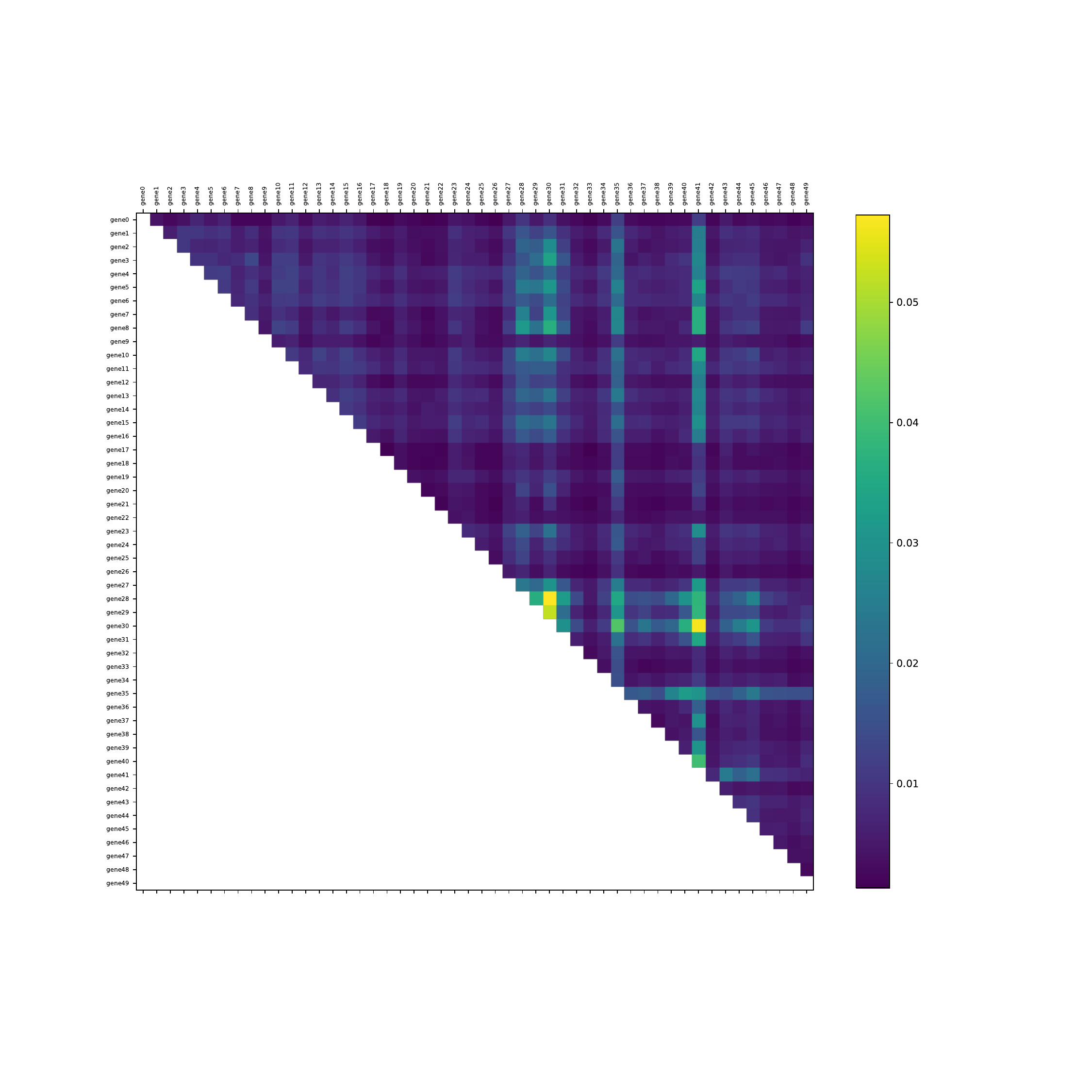}
    \includegraphics[width=0.325\textwidth,trim={4cm 2cm 4cm 4cm},clip]{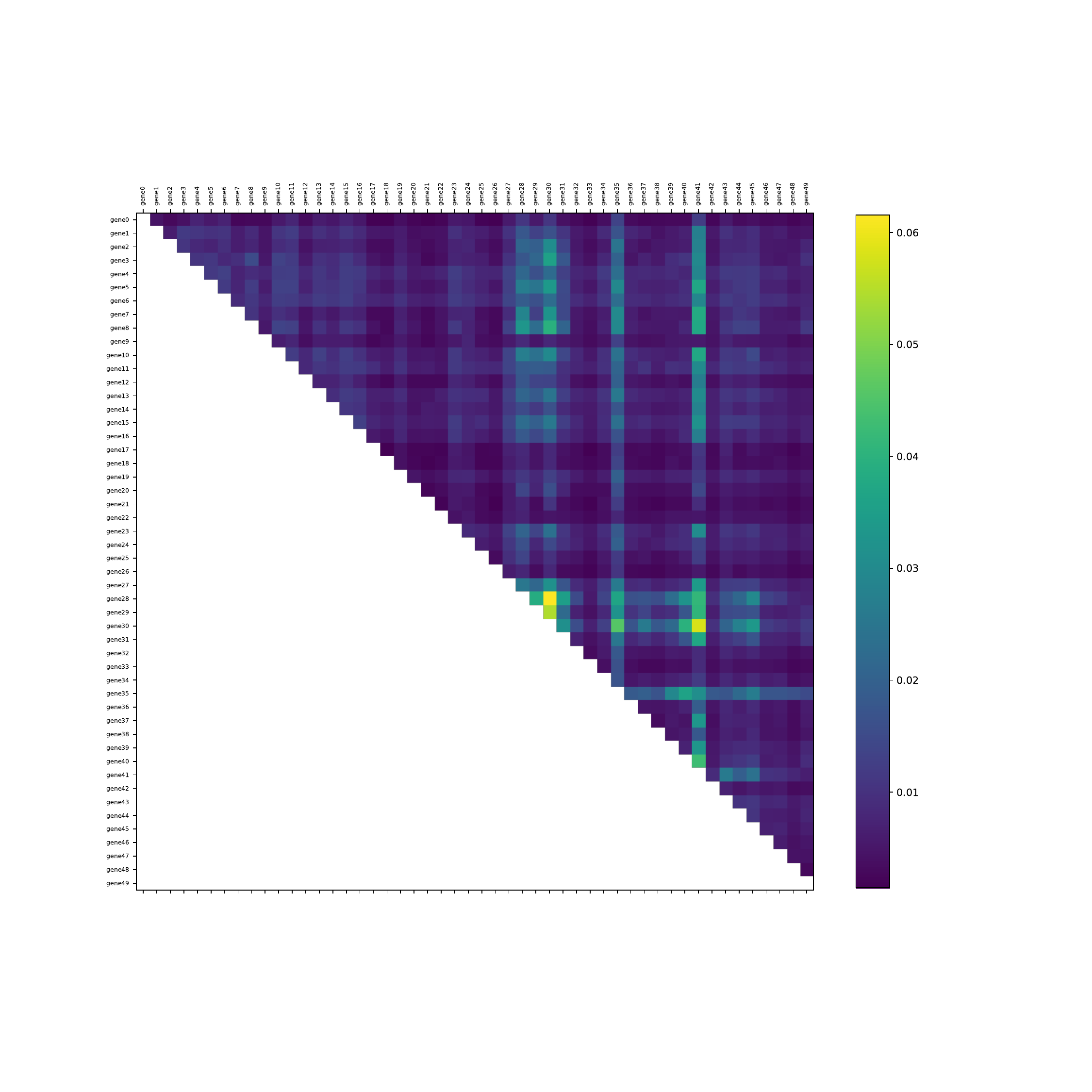}
    \includegraphics[width=0.325\textwidth,trim={4cm 2cm 4cm 4cm},clip]{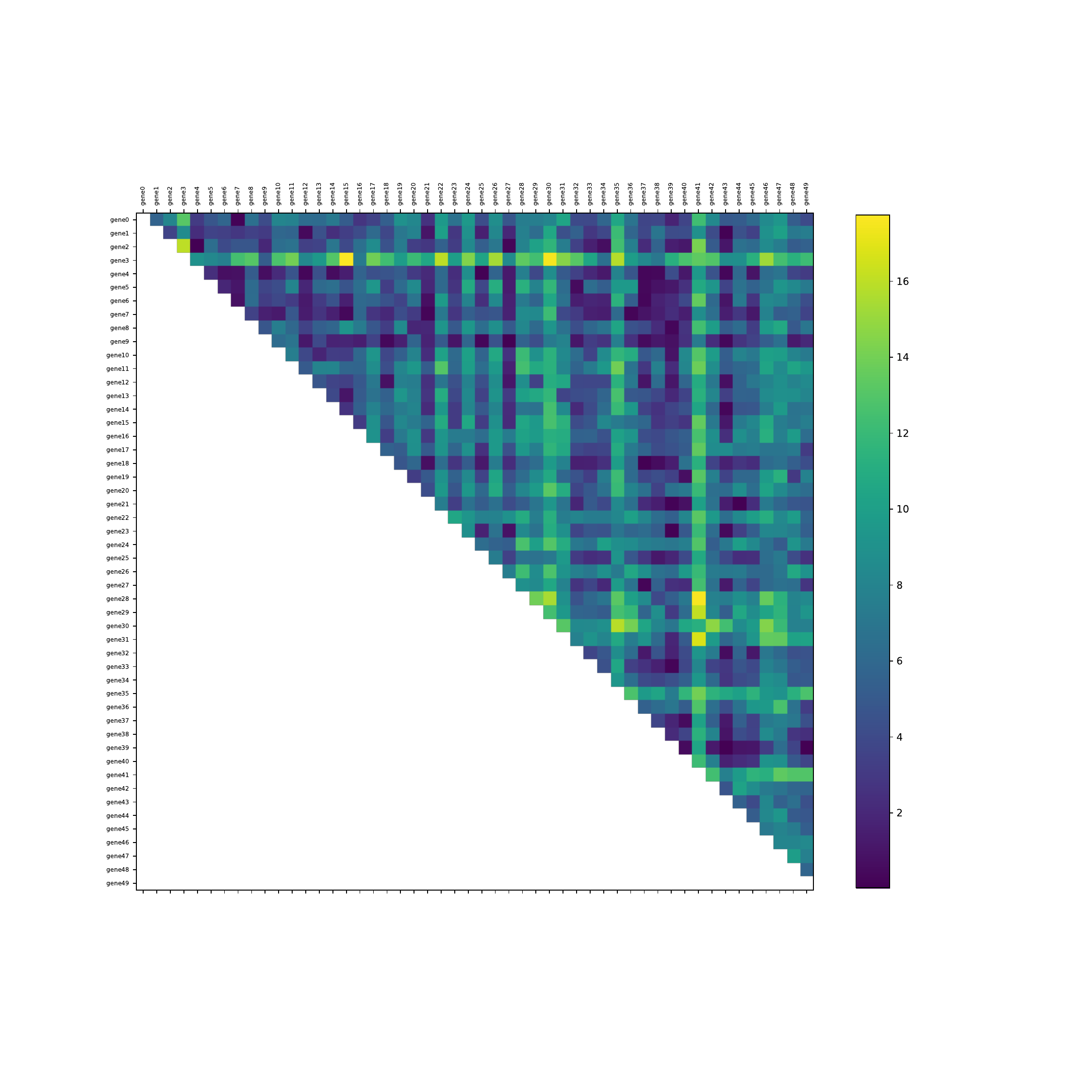}
    \vspace{-1em}
    \caption{Pairwise interaction scores from a selected set of $50$ genes using disjointedness test with Matern 2.5 kernel and RBF kernel (\emph{left} and \emph{middle}, respectively), and separability test (\emph{right}); brighter colors suggest stronger interactions. Genes from the same pathway are ordered adjacently. The associated pathways of each selected gene are described in \cref{tab:gene_indices_pathways} of \cref{apdx:genegene}.
    }
    \label{fig:realrwd}
\end{figure*}

We computed the test statistics using 1024-dimensional embeddings of the
original cell painting images extracted from a pre-trained masked autoencoder
\citep{he2022masked,kraus2023masked}.
\cref{fig:realrwd} shows the matrices of the test statistics for all
perturbation pairs; each entry represents the pairwise interaction scores.
Qualitatively, both the disjointedness statistics and separability statistics effectively
uncover plausible biological relationships.
Many genes in the apoptosis pathway (programmed cell death, e.g., gene 2--3: BAX, BCL2L1) and the proteasome (protein degradation e.g., gene 28--30: PSMA1, PSMB2, PSMD1) 
show high scores that are expected from synthetic lethal relationships. Synthetic lethality (SL) is a complex phenomenon in which simultaneous inactivation of specific gene combinations leads to cell death or extreme sickness, while individual perturbations have little effect. Apoptosis, controlling cell survival, is tightly regulated; disruption of the anti-apoptotic BCL-2 family (BCL2, BCL2L1, MCL1) \citep{kale_bcl-2_2018} interferes with critical barriers against cell death. Proteasome function is similarly regulated, as cells must maintain a delicate balance of different proteins \citep{Rousseau_2018}. Apoptosis and proteasome members are often found in SL screens in cancer cells \citep{li_2020, Ge_2024, Han_2017, cron_2013, Steckel_2012, Das_2020} because cancer genomes accumulate mutations that overcome weakened cellular buffering capabilities. 

As displayed in \cref{fig:realrwd}, the MMD-based statistics show strong signals of interactions with proteasome components; the proteasome helps control global protein levels so we would expect to see it interacting with many different pathways, some essential.
The separability score provides less clear patterns but highlights several gene pairs that are known or expected to interact physically or genetically, 
e.g., BCL2L1-MCL1 (gene 3-8) \citep{Shang_2020, Carter_2024}, BAX-BCL2L1 (gene 2-3) \citep{Lindqvist_2014}, BCL2L1-PSMD1 (gene 3-30) \citep{Craxton_2012}, and PSMB2-PSMD1 (gene 29-30) \citep{voutsadakis2017proteasome}.

\subsection{Interaction discovery with active matrix completion}

\begin{figure*}
    \centering
    \includegraphics[width=0.32\textwidth]{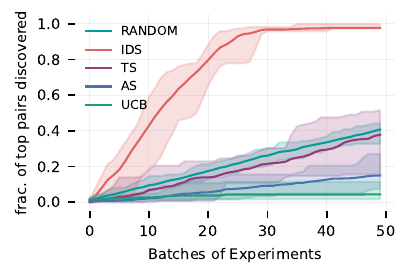}
    \includegraphics[width=0.32\textwidth]{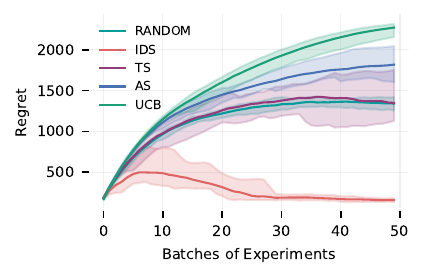}
    \includegraphics[width=0.32\textwidth]{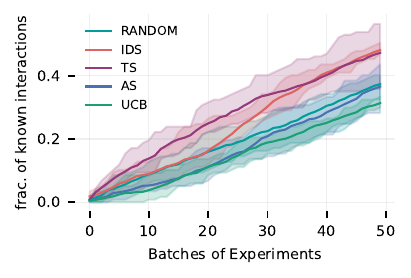}
    \caption{
    Empirical results on the active pairwise experiments selection for the gene-gene interaction detection example using MMD based test statistics.
    The solid lines represent the mean performance,
    whereas the shaded region represent all the runs (min-max). IDS (ours) outperforms all
    the baselines significantly in terms of top scoring pairs discovered as well as
    the regret. In terms of known interactions, IDS still outperforms the baselines
    by a small margin.
    }
    \label{fig:discovery_results}
\end{figure*}

Finally, we showcase how to use our test statistics to adaptively select the pairwise
experiment to run. As introduced in \cref{sec:active_learning}, the adaptive experimental design
is framed as an active matrix completion problem (\cref{alg:discovering_pairs}).
We applied \cref{alg:discovering_pairs} on the gene-gene interaction detection example using the MMD-based scores. 

\paragraph{Baselines.} To evaluate the effectiveness of our automated selection method, we compare against
selection with random policy, upper confidence bound
(UCB), Thompson sampling (TS) and uncertainty sampling (US). To instantiate UCB
in the discovery setting, we use the uncertainty from the low-rank matrix
posterior in place of the counts used in the standard multi-armed bandit
setting and mask the actions once they are selected. Similarly, in US we pick
pairs based solely on the uncertainty of the posterior. TS is instantiated in
the same way as IDS, but the pairs are selected to minimize only the instant
regret. Further details are discussed in \cref{apdx:expt}.

\paragraph{Evaluation metrics.}  We evaluate each approach using three different metrics given a budget of 50
rounds, each with a batch size of 10 resulting in a total of 500 experiments
covering $~50\%$ of all possible pairs of the gene set. 
First, we look at the fraction of the pairs with the top $5$ percentile of scores recovered by
the algorithm capturing the ability to explore the high scoring regions well.
Next, we also evaluate the regret of each algorithm with respect to an optimal
policy with access to the score matrix and acquires the highest scoring pairs
at each round. Finally, we evaluate the number of known biological relations
that appear in CORUM \citep{giurgiu2019corum}, StringDB
\citep{szklarczyk2021string}, Signor \citep{licata2020signor} and hu.MAP
\citep{drew2021hu} to see how many each method is able to recover.

\paragraph{Results.} \cref{fig:discovery_results} (and \cref{fig:discovery_results_rbf} in \cref{apdx:al}) illustrate the empirical results. We
observe that IDS discovers all the pairs with the top-5 percentile scores,
whereas all the baselines barely recover half of the top pairs. Even in terms
of the regret, IDS outperforms the baselines, with random performing the worst.
This demonstrates that IDS is able to exploit the low-rank structure in the
reward matrix effectively. Regarding known interactions discovered,
IDS and TS outperform other methods by a slight margin.
After 50 rounds,  we observe that IDS and TS achieve between $12$ and $15\%$ improvement over the other baselines in the number of known biological interactions.
The significant difference between the performance on the fraction
of top pairs and the number of known relations prompts questions about the
correlation between the prediction errors and the biological interactions.

\section{Discussion}
\label{sec:discussion}
This paper presented a method for efficiently detecting interactions between
perturbations. We present interactivity scores for both separability and disjointedness which both allow us to measure interactions between latent variables, and run
experiments only on pairs of perturbations for which we will fail to
compositionally generalize.
From an active learning perspective, disjointedness is powerful: when we can learn a good posterior over this test statistic, we effectively have a ``confidence score'' for whether two perturbations are likely to compose additively. This allows us to dramatically reduce the number of experiments we run by only experimenting where embeddings are unpredictable.
Separability gives a more intuitive notion of (in-)dependence in that we directly test for whether two perturbations interact in latent space. It will be interesting to explore to what extent this can be used to recover the target of interventions.
\paragraph{Limitations} While many known interactions scored highly according
to our test statistics, the overall correlation between known
interactions and this metric was relatively low. It is not clear whether this
is the result of a lack of specificity, or whether this is because we are in
fact discovering real relationships that are not known to biology. To test this
would require additional experimentation with orthogonal assays.
Additionally, the results for the separability tests depended on the quality of the KL estimator. We used the SMILE estimator which is somewhat sensitive to the choice of the clipping parameter, $\tau$. In our experiments on images, a default of  $\tau = 5$ seemed to work well, but if we want this test to be applicable across a variety of modalities, we need a more robust method for choosing this hyper-parameter.

\section{Acknowledgements}
\label{sec:ack}
We are extremely grateful for the many discussions with colleagues at Recursion and Valence Labs, and external collaborators that lead to this work. 
The original design of the benchmark dataset was designed by Marta Fay and the experiments were managed and run by Jordan Finnell, Brandon Mendivil, Kate Brown and Vicky Chen. Thank you to Nathan Lazar for his help in preparing the data and helpful discussions throughout the project.
The pairwise interaction testing section was strongly influenced by discussions with everyone at the Bellairs Causal Workshop; special thanks to Victor Veitch, Kartik Ahuja and Yixin Wang for their input. 
Finally, thank you to Cian Eastwood, Johnny Xi and Jana Osea for the helpful discussions and feedback on the work.

\bibliographystyle{abbrvnat}
\bibliography{references}

\clearpage
\appendix
\onecolumn

\addcontentsline{toc}{section}{Appendix} %
\part{Appendix} %
\parttoc %
\section{Proofs} \label{apdx:proof}

\bprfof{\cref{thm:lpdfscoreadd}}
By \cref{assump:causaldag,assump:diffeo}, we can apply change of variable formula to obtain the 
log density of $p_X(x|T_1, \dots, T_n)$:
\[
    \log p(x | T_1, \dots, T_n) = \sum_{l = 1}^L \log p_{z_l}( [g^{-1}(x)]_{l} |\pa(z_l)) + \log |\det \nabla g^{-1}(x))|. 
\]
Here $[\cdot]_l$ denote the projection operator that maps $z\in \scZ$ to the subspace on which $z_l$ lies, i.e., 
$\forall z \in \scZ, [z]_{l} = z_l$. 
Recall that we interpret $z$ as the concatenation of all latent factors $(z_1, \dots, z_L)$.
If $T_i$, $T_j$ are causally independent, perturbation $\delta_i, \delta_j$
will intervene different terms in the above summand. Without lose of generality, 
suppose $\delta_i$ and $\delta_j$ intervene $z_{l_i}$ and $z_{l_j}$ repsectively.
Then, 
\[
    \bar \ell_X(x|\delta_i) 
    & = \log p(x|\delta_i) - \log p(x|\delta_0) \\
    & = \log p_{z_{l_i}}([g^{-1}(x)]_{l_i}|\delta_i) - \log p_{z_{l_i}}([g^{-1}(x)]_{l_i}|\delta_0).
\]
where the equality follows from the modularity assumption that the intervention only affects $l_i$.

Similarly, we obtain that
\[
    \bar \ell_X(x|\delta_j) &= \log p_{z_{l_j}}([g^{-1}(x)]_{l_j}|\delta_j) - \log p_{z_{l_j}}([g^{-1}(x)]_{l_j}|\delta_0), \\
    \bar \ell_X(x|\delta_{ij}) & = \log p_{z_{l_i}}([g^{-1}(x)]_{l_i}|\delta_i) - \log p_{z_{l_i}}([g^{-1}(x)]_{l_i}|\delta_0) \\
                            &+ \log p_{z_{l_j}}([g^{-1}(x)]_{l_j}|\delta_j) - \log p_{z_{l_j}}([g^{-1}(x)]_{l_j}|\delta_0),
\]
which completes the proof.
\eprfof

\bprfof{\cref{thm:mixture}}
Provided with \cref{assump:condindep}, $\delta_i, \delta_j$ being disjoint (\cref{def:disjoint}) is equivalent to 
the same additivity in the intervened latent distributions, i.e., 
\[ \label{eq:latentexclusive}
    p_Z(z|\delta_{ij}) - p_Z(z|\delta_0) = p_Z(z|\delta_{i}) - p_Z(z|\delta_0) +  p_Z(z|\delta_{j}) - p_Z(z|\delta_0).
\]
Under the mixture model assumption \cref{assump:mixture}, if $T_i$, $T_j$ are causally independent, 
if $\ch(T_i) \cap \ch(T_j) = \emptyset$, 
perturbation $\delta_i, \delta_j$ will intervene distinct mixing components.
Without lose of generality, suppose $\delta_i$ and $\delta_j$ intervene $p_{Z_{l_i}}$ and $p_{Z_{l_j}}$ repsectively.
\[
    p_Z(z|\delta_i) - p_Z(z|\delta_0)  &= w_{l_i} \cdot (p_{Z_{l_i}}(z|\delta_i) - p_{Z_{l_i}}(z|\delta_0)), \\
    p_Z(z|\delta_j) - p_Z(z|\delta_0)  &= w_{l_j} \cdot (p_{Z_{l_j}}(z|\delta_j) - p_{Z_{l_j}}(z|\delta_0)), \\
    p_Z(z|\delta_{ij}) - p_Z(z|\delta_0)  &= w_{l_i} \cdot (p_{Z_{l_i}}(z|\delta_i) - p_{Z_{l_i}}(z|\delta_0)) 
                                          +w_{l_j} \cdot (p_{Z_{l_j}}(z|\delta_j) - p_{Z_{l_j}}(z|\delta_0)),
\]
which shows \cref{eq:latentexclusive} and hence completes the proof.
\eprfof

\section{Sample-based estimation of KL divergence} \label{apdx:klest}

Sample-based estimator of KL-divergence is a challenging task, particularly for high dimensional observations.
In this section, we present the estimation procedure we used in our experiments.
As mentioned in \cref{sec:experiments}, to estimate the KL-divergence,
\[\label{eq:kl}
    \kl{ p}{p_{i}} = \E\left[ \log \frac{p(X|\delta_{0})}{p(X|\delta_{i})}\right], \quad X \sim p(X|\delta_{0}),
\]
we used a two-step procedure:
\benum
\item We first estimate the log-density ratio 
    \[ \label{eq:ldr}
    \log \frac{p(x|\delta_{0})}{p(x|\delta_{i})}
    \]
    via a neural ratio estimator (NRE) based on a contrastive learning objective \citep[NRE]{hermans2020likelihood}.
\item Then, instead of taking naive Monte Carlo estimates of the KL-divergence, we adopt the SMILE estimator \citep{songunderstanding} using the learned log-density ratio.
\eenum
\cref{apdx:nre,apdx:smile} describe the two steps respectively.

\subsection{Contrastive neural ratio estimator} \label{apdx:nre}

There are various methods to learn the log-density ratio between a single pair
of distributions. For example, the optimal discriminator between two
distributions is related to their density ratios \citep[Proposition
1]{goodfellow2014generative}. However, a common structure in our applications
involves a large set of perturbations, and training a classifier for each
perturbation class (against the control group) would be very cumbersome and not
data efficient. Therefore, we consider a contrastive learning model
\citep[NRE]{hermans2020likelihood}, which trains a binary classifier to
distinguish the joint data distribution 
$p(x,c)$ from the product of the marginals 
$p(x)p(c)$, where $c$ denotes the perturbation class.

The training objective is as follows:
\[ \label{eq:nre}
    \theta, w \in \arg\min_{\theta, w} -\frac{1}{2B}\left[ \sum_{b = 1}^B \log( 1 - \text{Sigmoid}(f_{\theta, w}(x^{(b)}, c^{(b)})) +\sum_{b' = 1}^B \log(\text{Sigmoid}(f_{\theta, w}(x^{(b')}, c^{(b')})) \right],
\]
where $x^{(b)}, c^{(b)} \sim p(x)p(c)$ and $x^{(b')}, c^{(b')} \sim p(x, c)$, and 
\[
    f_{\theta, W}(x, c) = \text{Encoder}_\theta(x)^T W_c.
\]
Given infinite training samples and flexibility of the neural network $f$, the optimal $f^\star(x, c) = \log \frac{p(x, c)}{p(x)p(c)} = \log \frac{p(x|c)}{p(x)}$.
And we can then obtain \cref{eq:ldr} via $f^\star(x, \delta_i) - f^\star(x, \delta_0)$.

In our examples, we estimate our log-density ratios by training the NRE objective on all perturbation classes.

\subsection{Smoothed Mutual Information “Lower-bound” Estimator (SMILE)} \label{apdx:smile}
After obtaining the log-density ratio estimator for \cref{eq:ldr}, there are several options for estimating the KL-divergence. We provide a short review here on the various strategies \citep{ghimire2021reliable,belghazi2018mutual,hjelm2018learning,songunderstanding}, and explain why we opt for the SMILE estimator \citep{songunderstanding}.

The most straightforward estimates of the KL-divergence \cref{eq:kl} is by the Monte
Carlo estimates based on samples from $p(X|\delta_{0})$, i.e., 
\[
    \kl{P(X|\delta_i)}{P(X|\delta_0)} \approx \frac{1}{N}\sum_{i= 1}^N f(X_i), \quad \text{where } f(\cdot) \text{is the estimated} \log \frac{p(\cdot | \delta_i)}{p(\cdot|\delta_0)}. 
\]
However, it is noted that the variance of this estimator is often huge
\citep{songunderstanding, ghimire2021reliable}, making the estimated KL
unreliable in practice.

\citet{belghazi2018mutual} proposed to estimate the KL-divergence based on its Donsker-Varadhan representation \citep{donsker1983asymptotic}, i.e., 
\[
    \kl{p}{q} = \sup_{f: \Omega \to \reals} \E_p[f] - \log \E_q[\exp(f)], 
\]
where the supremum is taken over all functions $f$ such that the two expectations are finite.
Notice that the optimal $f$ is indeed achieved as the log-density ratio between $p$ and $q$.
In practice, one can parameterize $f$ using some neural network and maximize the above objective to approximate the KL \citep{belghazi2018mutual}. However, the stochastic gradient estimator of the above objective is generally biased \citep{belghazi2018mutual}, making the optimization less stable.
Therefore, it is also recommended to 
first learn the log-density ratio $\log \frac{p}{q}$ as $f$, and then estimate the KL-divergence \citep{hjelm2018learning,songunderstanding} via
\[
\text{KL}_\text{MILE}(f) = \E_p[f] - \log \E_q[\exp(f)],
\]
which technically gives a lower-bound of the KL divergence if the log-density ratio is not well-estimated. We would refer this estimator as the mutual information lower-bound estimator (MILE).
MILE is often shown to have better performance than the naive Monte Carlo estimator \citep{belghazi2018mutual,hjelm2018learning,songunderstanding}.

However, MILE also suffers from high variance issues \citep{songunderstanding} particularly when the learned $f$ has large values in the tail of $q$.
\citet{songunderstanding} proposed a smoothed version of MILE, named as SMILE, by clipping the learned log-density ratios $f$ between $-\tau$ and $\tau$, i.e., 
\[ \label{eq:smile}
    \text{KL}_\text{SMILE}(f, \tau) = \E_p[f] - \log \E_q[\text{clip}(\exp(f), \exp(-\tau), \exp(\tau))].
\]
$\text{KL}_\text{SMILE}(f, \tau)$ converges to $\text{KL}_\text{MILE}(f)$ as $\tau \to \infty$, but smaller $\tau$ significantly reduces the variance of the MILE estimator.

In our experiments, we use the SMILE estimator with the clipping parameter $\tau$ set to be $5$.

\section{MMD and kernel mean embedding} \label{apdx:mmd}

\subsection{Kernel mean embedding} \label{sec:kernelembedding}
We provide here a minimal overview of the kernel mean embedding of probability measures \citep{muandet2017kernel}.
Given a feature map $\phi:\scX \to \scF_\phi$, where $\scF_\phi$ is some Hilbert space (sometimes called the feature space).
This feature map $\phi$ defines a kernel: 
\[
    k_\phi:\scX\times\scX \to \reals, \qquad k_\phi(x, y) = \langle\phi(x), \phi(y) \rangle, 
\]
where $\langle\cdot, \cdot\rangle$ denotes the inner product of $\scF_\phi$.
Kernel $k_\phi$ defined this way induces a space of functions---$\scH_\phi$---from
$\scX$ to $\reals$, which is a \emph{reproducing kernel Hilbert space} (RKHS).
The name RKHS comes from a special property of $\scH_\phi$, called the \emph{producing property}:
\[
\forall f \in \scH_\phi, \qquad \langle f, \phi(x) \rangle_{\scH_\phi} = f(x).
\]
We abuse the notation here by interpreting $\phi(x)$ as a function in
$\scH_\phi$ instead of a real value. 
A special instance of the reproducing property is that 
\[
\langle \phi(x), \phi(y)\rangle = k_\phi(x, y).
\]
Precisely,  we are using the following representation of $\phi(x)$:
\[
\phi(x):\scX \to \scH_\phi, \qquad \phi(x)(\cdot) = k_\phi(x, \cdot) \in \scH_\phi.
\]
Here $k_\phi(x, \cdot) \in \scH_\phi$ is guaranteed by the definition of $\scH_\phi$.

In what follows, we use $\scM^1_+(\scX)$ to denote the space of probability measures over $\scX$.
We can embed any probability measure to the RKHS.
\bdefn[Kernel mean embedding of probability measure]\label{def:kernelmeanemdprob}
The \emph{kernel mean embedding} of a probability measure $\P \in \scH_\phi$ is defined via the following mapping:
\[
    \mu_\phi: \scM^1_+(\scX) \to \scH_\phi, \qquad \mu_\phi(\P)(\cdot) = \int_{\scX} \phi(x)(\cdot) \P(\d x) = \E_{X\sim \P}\left[\phi(X)\right](\cdot).
\]
\edefn
We denote the kernel mean embedding of $\P$ by $\mu_\phi(\P)$. 

\bprop[c.f Eq. (3.29) of \citet{muandet2017kernel}] \label{prop:mmdl2norm}
For all $\P, \Q \in \scM^1_+(\scX)$,
\[\label{eq:mmdl2norm}
    \MMD_{k_\phi}(\P, \Q) = \left\|\E_{X\sim \P}\left[\phi(X)\right] - \E_{Y\sim \Q}\left[\phi(Y)\right]\right\| = \| \mu_\phi(\P) - \mu_\phi(\Q) \| .
\]
\eprop
\bprfof{\cref{prop:mmdl2norm}}
The second equality of \cref{eq:mmdl2norm} follows directly from the mapping defined in \cref{def:kernelmeanemdprob}.
We then focus on proving the first equality.
\[
&\MMD_{k_\phi}(\sdM_1, \sdM_2) \\
  &= \sup_{f \in \scH_\phi: \|f\|\leq 1} \E_{X\sim \P}[f(X)] - \E_{Y\sim \Q}[f(Y)]\\
  &= \sup_{f \in \scH_\phi : \lVert f \rVert \le 1} \E_{X \sim \P}[\langle f, \phi(X)\rangle] - \E_{Y \sim \Q}[\langle f, \phi(Y) \rangle] \quad \text{(reproducing property)}\\
  &= \sup_{f \in \scH_\phi : \lVert f \rVert \le 1} \langle f, \E_{X \sim \P}[\phi(X)]\rangle - \langle f, \E_{Y \sim \Q}[\phi(Y)]\rangle \quad \text{(linearity of inner product)}\\
  &= \sup_{f \in \scH_\phi : \lVert f \rVert \le 1} \langle f, \E_{X \sim \P}[\phi(X)] -\E_{Y \sim \Q}[\phi(Y)]\rangle  \quad \text{(linearity of inner product)}\\
\]
Then the proof is completed by the observation that the inner product is maximized when  
\[
f = \left\{\E_{X \sim \P}[\phi(X)] -\E_{Y \sim \Q}[\phi(Y)] \right\}/\left\|\E_{X \sim \P}[\phi(X)] -\E_{Y \sim\Q}[\phi(Y)] \right\|.
\]
\eprfof
\cref{prop:mmdl2norm} essentially says that MMD between two distributions
is indeed the distance of mean embeddings of features.
It also tells us that $\MMD_{k_\phi}(\P, \Q)= 0$ if and only if $\mu_\phi(\P) = \mu_\phi(\Q)$.
To be able to separate any two distributions via MMD, the kernel mean embedding must be an injective map, 
in which case the feature map induces a \emph{characteristic kernel}.
\bdefn[Characteristic kernel]
$k$ is said to be characteristic on $\scM^1_+(\scX)$ if the kernel mean embedding,
\[
    \mu_k: \scM^1_+(\scX) \to \scH_k, \qquad \mu_k(\P)(\cdot) = \int_{\scX} \phi(x)(\cdot) \P(\d x),
\]
is injective. In other words, $k$ is a characteristic kernel if and only if
\[\label{eq:characteristickernel}
\mu_k(\P, \Q) = 0 \Longleftrightarrow \mu_k(\P) = \mu_k(\Q)  \Longleftrightarrow \P = \Q.
\]
\edefn
Here we change the subscript of $\mu$, $\scH$ from the feature map to the kernel, 
because from now on we do not necessarily work on explicit choice of feature maps;
many characteristic kernels do not have tractable feature maps (and mostly infinite dimensional). 
\cref{eq:characteristickernel} states that MMD is a metric on $\scM^1_+(\scX)$ if a characteristic kernel is used; otherwise, distinct distributions with the same kernel mean embedding cannot be separated by MMD.
Examples of characteristic kernels on $\scM^1_+(\reals^d)$ are Gaussian kernels, Laplacian kernels, and the family of Matérn kernels.
We refer readers to \citet{sriperumbudur2011universality} for a comprehensive
survey of the characteristic kernels.

Provided with a specified kernel, and samples from $\P, \Q$, one can obtain an unbiased estimate of the squared population MMD; 
see \citet[Eq. (3)]{gretton2012kernel} for the detailed expression of the estimator.
An interesting property of the estimates of MMD is that the convergence is
dimension independent \citep[Theorem 7]{gretton2012kernel}, 
although the penalization of the dimension in the context of kernel two sample
test lies in the reduction of power \citep{ramdas2015decreasing}.

\subsection{Compare distributions via a fixed feature map $h$} \label{apdx:chooseh}

In some cases, one can compare $\P, \Q$ by assessing on a fixed test function $h$. 
For example, the L2 norm between the expectation of $h$, i.e., 
\[ \label{eq:l2rewardh}
    \|\E_{X\sim \P}[h(X)] - \E_{Y\sim \Q}[h(Y)]\|_2
\]
can be a crude measure on how different $\P, \Q$. According to \cref{prop:mmdl2norm},
\[
    \|\E_{X\sim \P}[h(X)] - \E_{Y\sim \Q}[h(Y)]\|_2 = \mathrm{MMD}_{k_h}(\P, \Q).
\]
The obvious limitation with a fixed choice of $h$ is that it's typically not characteristic for all distributions, i.e., 
\[
    \|\E_{X\sim \P}[h(X)] - \E_{Y\sim \Q}[h(Y)]\|_2 = 0 \centernot \implies \P = \Q.
\]
However, in practice, if the choice of $h$ is sufficient to identify discriminate the
collection of distributions that we care about, it is convenient to just examine \cref{eq:l2rewardh}.

In our specific application, interaction detection, we find the embedding vector that are
constructed from the final hidden layer of a classifier works reliably. 
We assume that this classifier is trained optimally such that, $\P(\pert_i | \observed) =
\sigma(w_i^\top h(\observed))$, where $\sigma :=
\frac{\exp(x_i)}{\sum_j\exp(x_j)}$ is the softmax function, and that there are
sufficiently diverse labels to ensure that this representation is identified up
to a linear transformation; see \citet{roeder21a} for details. Given a trained
classifier, 
$\sbh_i := \E[h(\observed)|\pert_i] - \E[h(\observed)|\delta_0]$ 
is then the average of the embeddings associated with a particular knockout centered around the control wells.
We find in our empirical experiments that the metric $\|\sbh_{ij} - \sbh_i - \sbh_j\|_2$ 
works well to identify highly interacting gene pairs.

We have not obtained theoretical arguments on justifying the use of optimal discriminator as the choice of test functions;
we left this for future development.

\section{Experiment details} \label{apdx:expt}

For all separability tests, we trained the NRE model on all perturbation
classes and then obtained the log-density ratios for each pair as introduced in \cref{apdx:nre}.
For the NRE training, we evaluated multiple model architectures and optimizer
step sizes, selecting the hyperparameter combination that yielded the best
training accuracy. The selected model architectures and optimizer step sizes
for each example are reported in the corresponding sections. We checkpointed
the best model at the optimal training accuracy for further inference to obtain
density ratio estimates and KL estimates. To optimize our models, we used ADAM \citep{adam} with default
hyperparameter settings.

All our experiments were run on NVIDIA H100 GPUs.

\subsection{Synthetic examples}
\label{apdx:sync}
For all the synthetic examples, we generated 20,000 \iid samples for each
perturbation class, including both single and double perturbations. The
detailed data generation process for each example is provided below.

\subsubsection{Synthetic tabular for separability test}     

The separability scores in this example are computed using two different
KL-divergence estimators: the simple KNN-based estimator and the NRE-based
estimation procedure described in \cref{apdx:klest}. We used a 3-layer MLP with
ReLU activation (hidden dimensions of 128 and 64) as the encoder for the NRE
density ratio estimator. The NRE model was trained using the ADAM optimizer
with a step size of 0.005 for 500 epochs and a batch size of 1024.

\paragraph{Latent distribution} 
The latent variable $Z$ consists of 3 independent one-dimensional variables $P_1, P_2, P_3$, i.e., 
\[
P_Z(z_1, z_2, z_3) = P_1(z_1) \cdot P_2(z_2) \cdot P_3(z_3),
\]
for which we consider 4 single perturbations in total, labelled as $A, B, C, D$ respectively, resulting $6$ pairwise perturbations.
Perturbations are applied by changing the distribution of one or more of latent variables.
In our setting, perturbations are classified as separable or inseparable. 
Double perturbations of separable ones will intervene in two distinct latent
variables, while double perturbations of inseparable ones will intervene in the same latent variable.
For each node, the control (unperturbed) distribution is $\Norm(0, 1)$, while the perturbed one---whether it's single perturbed or doubly perturbed---becomes $\Norm(3, 1)$.

The association between the perturbation class and corresponding intervened latent variable(s) is described as follows:
\[\label{eq:syncdag}
\begin{aligned}
    P_1 &\sim 
    \begin{cases} 
        \text{Normal}(0,1) & \text{if unperturbed} \\ 
        \text{Normal}(3, 1) & \text{if perturbed by (A) or/and (B)} 
    \end{cases} \\
    P_2 &\sim 
    \begin{cases} 
        \text{Normal}(0, 1) & \text{if unperturbed} \\ 
        \text{Normal}(3, 1) & \text{if perturbed by (B)} 
    \end{cases} \\
    P_3  &\sim 
    \begin{cases} 
        \text{Normal}(0, 1) & \text{if unperturbed} \\ 
        \text{Normal}(3, 1) & \text{if perturbed by (C) or/and (D)} 
    \end{cases}
\end{aligned}
\]
As per \cref{eq:syncdag}, the only two inseparable pairs are A-B (both intervening $P_1$) and C-D (both intervening $P_3$).

\paragraph{Sampling process} 
To generate observations for each perturbation class, we first generate latent
samples from $P_Z$ based on the \cref{eq:syncdag}, and then transform the
latent samples via a diffeomorphism $g(\cdot)$. In this example, we chose
$g(\cdot)$ as a randomly initialized 7-layer Multi layer perceptron (MLP) with
LeakyReLU activations.

\subsubsection{Synthetic image example for separability test}

\begin{figure*}
    \centering
    \includegraphics[width=0.45\textwidth]{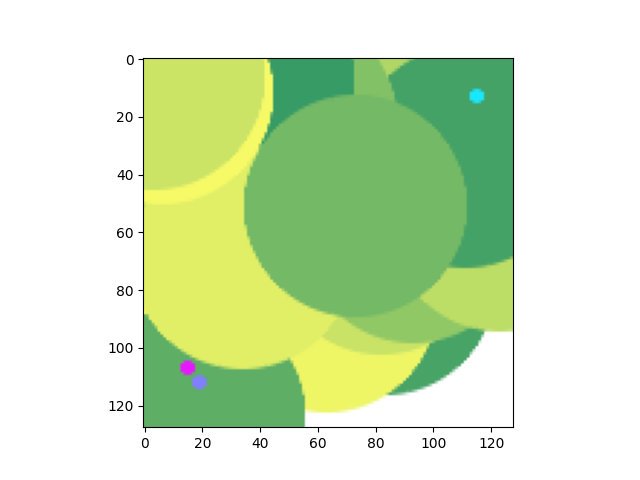}
    \includegraphics[width=0.45\textwidth]{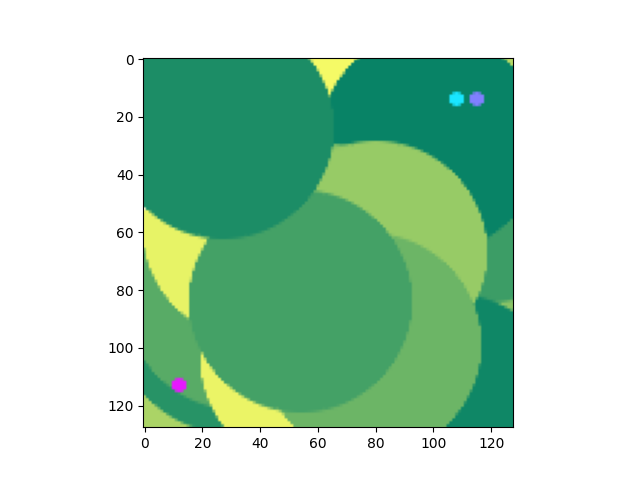}
    \caption{
        Example images of the interventional image data. 
        3 small balls with blue, purple, and red colors respectively are the targeted objects, whose locations are intervened by perturbations.
    Large balls with green and yellow colors form the background.
    }
    \label{fig:exampleimage}
\end{figure*}

The synthetic images consist of three objects (three small colored balls) in
different locations and with various backgrounds. The positions of the objects
are encoded in a 3-dimensional latent variable $Z$, which follows the
identical DAG structure described in \cref{eq:syncdag}. Each coordinate of
$Z$ corresponds to the location distribution of an object, determining the
distribution of both the $x$ and $y$ coordinates of the object. Thus, a
perturbation affecting the location distribution of one object will intervene
in the distribution of both coordinates of that object. Background distortion
is controlled by random noise. Scenes are generated using a rendering engine
from PyGame, denoted as $g(\cdot)$. Example images are provided in
\cref{fig:exampleimage}.

For the NRE model, we used a 5-layer convolutional network (with channels 32,
54, 128, 256, and 512) featuring batch normalization and leaky ReLU activations
as the encoder. The model was trained using the ADAM optimizer with a step size
of 0.0002 for 200 epochs and a batch size of 1024.

\subsubsection{Synthetic tabular example for disjointedness test}

The latent distribution is set to be a 6-component mixture distribution, 
\[
    P_Z(z) = \frac{1}{8} P_0(z) +  \frac{1}{8} P_1(z) +  \frac{1}{8} P_2(z) +  \frac{1}{8}P_3(z) +  \frac{1}{4} P_4(z) + \frac{1}{4} P_5(z).
\]
We define 7 single perturbations, labeled as A, B, C, D, E, F, G, respectively. 
The control distributions (unperturbed distribution) of all mixture component are $\Norm(0, 1)$.
Perturbations are applied by changing the distribution of one or more of the mixture components. 
E.g., if perturbation $D$ is applied, $P_4 \sim \text{Normal}(0, 5)$. 
In our setting, perturbations are classified as independent ones and interacting ones, 
where double perturbation of independent ones will intervene two distinct mixture components, 
while double perturbation of interacting ones will intervene the same mixture component.
The association of the perturbations and the corresponding component gets intervened is provided as follows:

The independent group consists of the mixture components \( P_0, P_1, P_2, P_3\):
\[
\begin{aligned}
    P_0 &\sim \text{Normal}(0, 1) &  \text{(not intervened)}\\
    P_1 &\sim \text{Normal}(5, 1) & \text{(A)}  \\
    P_2 &\sim \text{Normal}(10, 1) & \text{(B)} \\
    P_3 &\sim \text{Normal}(-5, 5) & \text{(C)} 
\end{aligned}
\]
The interacting group consists of the variables \( P_4, P_5 \):
\[
\begin{aligned}
    P_4 &\sim 
    \begin{cases} 
        \text{Normal}(0, 5) & \text{(D)} \\ 
        \text{Normal}(-10, 5) & \text{(E)} \\ 
        \text{Normal}(10, 5) & \text{(DE)} 
    \end{cases} \\
    P_5 &\sim 
    \begin{cases} 
        \text{Cauchy}(15, 1) & \text{(F)} \\ 
        \text{Cauchy}(-15, 1) & \text{(G)} \\ 
        \text{Cauchy}(20, 1) & \text{(FG)} 
    \end{cases}
\end{aligned}
\]

\paragraph{Sampling process}

To generate latent samples, we draw from the specified distributions and the constructed mixture model. Let \( n \) be the number of samples and \( d \) the dimension. The samples are generated as follows:
\[
\begin{aligned}
    &\text{if no intervention:} \quad \mathbf{X} \sim P_0 \\
    &\text{if intervention is specified:} \quad (\mathbf{P}_s, \mathbf{w}_s) = \text{build\_mixture}(\text{intervene}) \\
    &\text{draw categorical samples:} \quad \mathbf{z} \sim \text{Categorical}(\mathbf{w}_s) \\
    &\text{generate samples from each component:} \quad \mathbf{X}_i \sim \mathbf{P}_s[i]
\end{aligned}
\]
We generated latents for all 7 single perturbations, and all double perturbations, and 
then obtained the observations, which are used to perform tests on,  by mapping the latent samples through a deterministic function $g(\cdot)$. 
We chose $g(\cdot)$ as a randomly initialized 10-layer MLP with LeakyReLU activations.

\subsection{Real data examples}

\subsubsection{Gene-gene interactions} \label{apdx:genegene}

For the separability test, we used a 3-layer MLP with ReLU activation (hidden
dimensions of 2048 and 256) as the encoder for the NRE density ratio estimator.
We trained the NRE model using the ADAM optimizer with a step size of $0.0001$
for 2500 epochs and a batch size of 16384.

\cref{tab:gene_indices_pathways} describes the corresponding pathways for each selected gene. 

\begin{table}[]
\centering
\caption{List of gene indices and their associated pathways}
\label{tab:gene_indices_pathways}
\begin{tabular}{ll}
\textbf{Gene Index} & \textbf{Pathway} \\ 
[0.5ex]
\hline\hline
gene0  & Amino acid sensing (mTOR pathway) \\ 
\hline
gene1  & \multirow{9}{5em}{Apoptosis} \\ 
gene2  & \\ 
gene3  & \\ 
gene4  & \\ 
gene5  & \\ 
gene6  & \\ 
gene7  & \\ 
gene8  & \\ 
gene9  & \\ 
\hline
gene10 & \multirow{10}{5em}{Autophagy} \\ 
gene11 & \\ 
gene12 & \\ 
gene13 & \\ 
gene14 & \\ 
gene15 & \\ 
gene16 & \\ 
gene17 & \\ 
gene18 & \\ 
gene19 & \\ 
\hline
gene20 & \multirow{3}{10em}{ERAD (protein folding)} \\ 
gene21 &  \\ 
gene22 &  \\ 
\hline
gene23 & Integrated Stress Response \\ 
\hline
gene24 & \multirow{3}{5em}{Microtubule} \\ 
gene25 &  \\ 
gene26 & \\ 
\hline
gene27 & PI3K-Akt signaling \\ 
\hline
gene28 & \multirow{3}{5em}{Proteasome} \\ 
gene29 &  \\ 
gene30 &  \\ 
\hline
gene31 & Protein translation \\ 
\hline
gene32 & Protein translation (mTOR pathway) \\ 
\hline
gene33 & \multirow{3}{5em}{Ribosome} \\ 
gene34 &  \\ 
gene35 &  \\ 
\hline
gene36 & Transcriptional regulation \\ 
\hline
gene37 & \multirow{3}{10em}{UPR (protein folding)} \\ 
gene38 &  \\ 
gene39 &  \\ 
gene40 &  \\ 
gene41 &  \\ 
gene42 &  \\ 
\hline
gene43 & \multirow{3}{10em}{mTOR signaling }\\ 
gene44 &  \\ 
gene45 &  \\ 
gene46 &  \\ 
gene47 &  \\ 
gene48 &  \\ 
\hline
gene49 & p53 signaling \\ 
\end{tabular}
\end{table}

\subsubsection{Guide-guide interactions}
The single-cell painting images are derived from multi-cell images, with each
single-cell nucleus centered within a \(32 \times 32\) pixel box. The encoder
of the NRE model maps single-cell images of shape \((6, 32, 32)\) into a
128-dimensional feature vector. It consists of three convolutional blocks, each
comprising a Conv2D layer with a 3x3 kernel, BatchNorm2D, ReLU activation, and
MaxPool2D, progressively increasing the number of channels from 6 to 32, 64,
and 128 while halving the spatial dimensions at each max-pooling step. After
the convolutional layers, the output tensor of shape \((128, 4, 4)\) is
flattened to \((2048)\) and passed through two fully connected blocks, each
with a Linear layer, ReLU activation, and Dropout (with a dropout rate of 0.3),
transforming the feature size from 2048 to 256 and finally to 128. We train the
NRE model with the ADAM optimizer, using a step size of $0.00005$ for 5000
epochs and a batch size of 2048.

\subsection{Active learning} \label{apdx:al}

To obtain samples from the posterior distribution over the low-rank reward
matrix (with rank $m$), we use stochastic variational
inference~\citep{wingate2013automated,ranganath2014black}, and specifically the
implementation from
\texttt{numpyro}~\citep{jax2018github,bingham2019pyro,phan2019composable}. We
train the variational posterior for $5000$ epochs with a learning rate of $0.01$ using the Adam optimizer. We then generate $k$ samples from the posterior. 
For each algorithm, we run a sweep over all a set of hyperparameters. 
We then pick the best hyperparameters and run the experiment over 10 different seeds to get the final results. 
For IDS, we tune $m\in \{3, 5, 7, 10, 12\}$, $\lambda \in \{2, 3, 4, 5\}$ and $k\in \{500, 750, 1000, 1500\}$. 
For TS and US, we tune $m\in \{3, 5, 7, 10, 12\}$ and $k\in \{500, 750, 1000, 1500\}$. 
For UCB we tune $m\in \{3, 5, 7, 10, 12\}$, $\beta \in \{0.01, 0.1, 0.2, 0.5, 1, 2, 5\}$ and $k\in \{500, 750, 1000, 1500\}$, where $\beta$ controls the exploration.

\begin{figure*}
    \centering
    \includegraphics[width=0.32\textwidth]{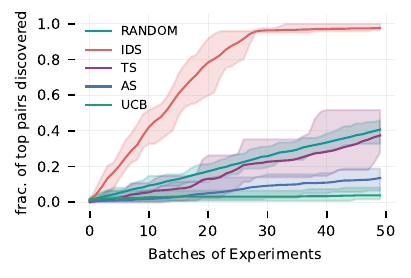}
    \includegraphics[width=0.32\textwidth]{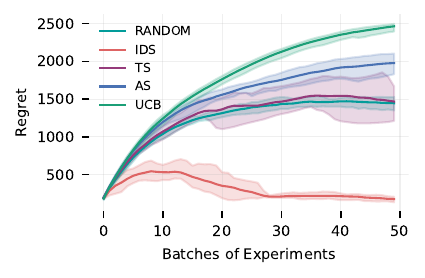}
    \includegraphics[width=0.32\textwidth]{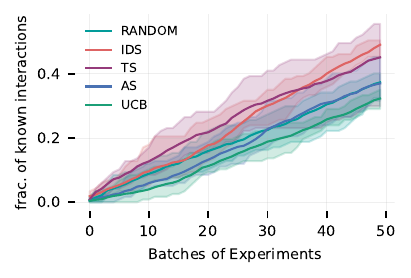}
    \caption{
    Empirical results on the active pairwise experiments selection for the gene-gene interaction detection example using MMD-based (using RBF kernel) test statistics.
    The solid lines represent the mean performance,
    whereas the shaded region represent all the runs (min-max). IDS (ours) outperforms all
    the baselines significantly in terms of top scoring pairs discovered as well as
    the regret. In terms of known interactions, IDS still outperforms the baselines
    by a small margin.
    }
    \label{fig:discovery_results_rbf}
\end{figure*}

\end{document}